# Empirical Investigation into Configuring Echo State Networks for Representative Benchmark Problem Domains

Brooke R. Weborg and Gursel Serpen
Electrical Engineering and Computer Science Department, The University of Toledo, Toledo, Ohio 43606, USA

**Abstract**

This paper examines Echo State Network, a reservoir computer, performance using four different benchmark problems, then proposes heuristics or rules of thumb for configuring the architecture, as well as the selection of parameters and their values, which are applicable to problems within the same domain, to help serve to fill the 'experience gap' needed by those entering this field of study. The influence of various parameter selections and their value adjustments, as well as architectural changes made to an Echo State Network, a powerful recurrent neural network configured as a reservoir computer, can be challenging to fully comprehend without experience in the field, and even some hyperparameter optimization algorithms may have difficulty adjusting parameter values without proper manual selections made first. Therefore, it is imperative to understand the effects of parameters and their value selection on Echo State Network architecture performance for a successful build. Thus, to address the requirement for an extensive background in Echo State Network architecture, as well as examine how Echo State Network performance is affected with respect to variations in architecture, design, and parameter selection and values, a series of benchmark tasks representing different problem domains, including time series prediction, pattern generation, chaotic system prediction, and time series classification, were modeled and experimented on to show the impact on the performance of Echo State Network.

# Introduction:

The influence of various parameter selections and value adjustments as well as architectural design changes made to an Echo State Network (ESN), a powerful recurrent neural network, on its performance is evaluated [1]. The study considered data in several problem domains including pattern classification, time series forecasting, pattern generation, adaptive filtering and control, system approximation, and short-term memory[2],[3],[4]. As a version of recurrent neural networks (RNN) also known as reservoir computers, ESN can capture trends in temporal data. Unlike other RNNs though, the ESN features a randomly generated hidden layer (also known as the reservoir) and only train the weights for the output layer[1],[5]. This approach is less computationally expensive than that for an RNN trained with gradient descent and also avoids the vanishing gradient issue [6],[7]. A study also suggested that ESN has a weak dependence on the size of the training set, compared to a feed-forward artificial neural network and a long short-term memory network (LSTM) [7].

Although ESN has the potential to address some of the training related difficulties or issues of RNNs trained with the gradient descent-based algorithm, they still face training related challenges of their own. One major issue is the fact that a randomly generated reservoir cannot be optimal and there is a need for some method to create an optimal reservoir [6],[7],[8],[9]. Some of the other specifics of issues for ESNs are defined as follows: Properties of the reservoir are poorly understood [10]; specification of the reservoir and input connections is an ad hoc process [10]; different applications of problem domains may require different reservoir configurations [11]; simply imposing a constraint on spectral radius value of the reservoir matrix may not be adequate to properly set the reservoir parameters [11]; and the random determination of the connectivity and weight structure of the reservoir is not likely to be optimal [11] [12].

Given these points, experience is a key factor to understanding parameter selection, regardless of if one is making manual selections or applying a hyperparameter optimization algorithm. Thus, this research aims to examine ESN using four different benchmark problems, such that they are easily reproducible, define guidelines that are applicable to problems within the same domain, and serve to fill the 'experience gap' needed to employ the ESN for a specific problem domain. Several domains were identified as good candidates from the literature. The problems and their associated domains which this study will analyz include the following: NARMA time series prediction task[13], Lazy Figure-8 pattern generation task [2], Mackey-Glass 17 chaotic system [14], and the Isolated Digit time series classification task [15], [16].

The NARMA benchmark evaluates the memory capability of ESN, as NARMA-10 is dependent on the previous ten timesteps [9] as well as the ability of the ESN to capture the nonlinearity of the system model. The Lazy Figure-8 benchmark specifically exploits the stability issues that arise when output feedback is required, which is mandatory in all pattern generating ESN models [2]. The Mackey-Glass sequence a popular chaotic system benchmark with a variety of reservoir computing studies using it ([17],[12],[18],[19]). The fourth benchmark problem, namely the free-spoken digit dataset, is a time series classification task. It was chosen based on its similarity to the datasets referenced in[8],[12], and [2]. This dataset will serve to demonstrate the modeling capabilities of ESN on high-dimensional data [12] as well as facilitating testing of parameter and architectural variations for the time series classification domain.

**Echo State Networks for Benchmark Problems**

Figure 1 shows the basic structure of an echo state recurrent neural network, which is a reservoir computer, with three layers of neurons: the input layer, the hidden layer, and the output layer. The internal structure of the hidden layer differs from other neural networks in that there are recurrent connections or topological cycles [5],[6]. Here, the discrete-time Echo State RNN presented in [5] and [20] will be used for discussion, with notation more similar to that presented in [21].

This model has $K$-dimensional input, $N$ neurons within the hidden layer, and $L$-dimensional output. The input is represented with the column vector $u(t)$, where $t$ is a specific point in time with $K$ features, and the weighted connections from the input to the hidden layer are represented by matrix $W_{in} \in \mathbb{R}^{N \times K}$. The hidden layer is represented by state vector $x(t)$ and the weighted connections between the $N$ neurons in the hidden layer represented by $W \in \mathbb{R}^{N \times N}$. Lastly, the output layer is represented by column vector $y(t)$, where $t$ is a specific point in time with $L$ targets, and the weighted connections from the hidden layer and input layer to the output layer are represented by $W_{out} \in \mathbb{R}^{L \times (N+K)}$. It is relevant to note that the input layer connection to the output layer is notated here but the connection is optional, as seen in Figure 1. Optionally, an RNN can have connections from the output layer that feed back into the hidden layer: these feedback connections are represented by $W_{fb} \in \mathbb{R}^{N \times L}$.

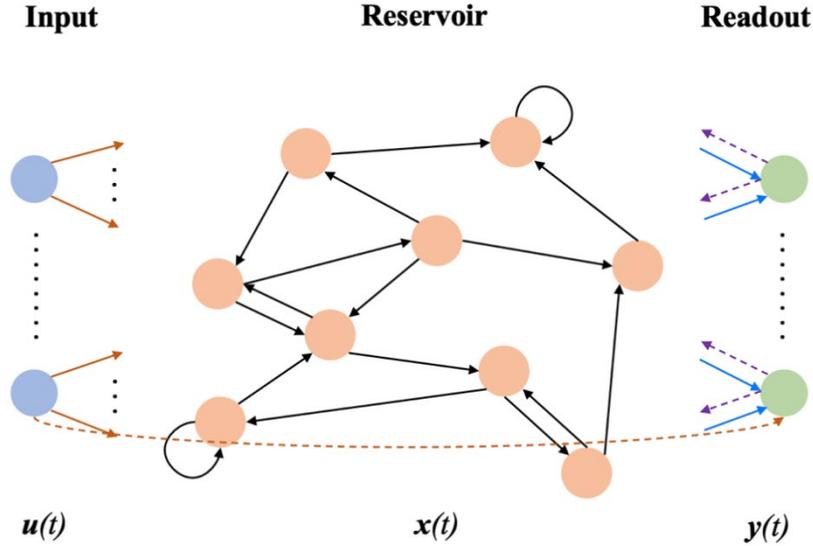

**Figure 1:** A recurrent neural network known as an echo state network. Blue circles represent the input layer $u(t)$, orange circles represent the hidden layer $x(t)$, and the green circles represent the output layer $y(t)$. The solid dark orange arrows represent $W_{in}$, the solid black arrows represent $W$, and the solid blue lines represent $W_{out}$. The dashed dark orange arrow represents an optional connection from the input to the output while the dashed purple arrows represent optional connections from the output back into the reservoir, $W_{fb}$.

For the discrete-time version of the Echo State RNN presented in Figure 1, the activation of internal units is updated by [1], [5],[6],[20]:

$$x(t + 1) = f(W_{in}u(t + 1) + Wx(t) + W_{fb}y(t)) \qquad (1)$$

where $f$ is an activation function of choice applied component-wise (usually $f = tanh$). The output is then calculated by the following [1],[5],[6],[21]:

$$y(t + 1) = g(W_{out}[u(t + 1), x(t + 1)]) \qquad (2)$$

where $g$ is an activation function that is applied element wise to its argument. This activation function is commonly selected as either $g = tanh$ or $g = 1$ (identity). Note that $[u(t + 1), x(t + 1)]$ is the concatenation of the input vector and the state vector; this is representative of an optional connection between the input layer and output layer, illustrated in Figure 1.

The reservoir, $W$, is comprised of various directed connections between the neurons. To generate $W$, one must start by randomly generating an $N \times N$ directional adjacency matrix with some defined density, $0 \leq d_W \leq 1$. This adjacency matrix represents where connections exist between any two neurons. For every existing connection in the adjacency matrix, which is represented by a value of "1" in the corresponding row and column of the matrix, its weight value is represented by a randomly selected number from the uniform distribution [-1,1]; this distribution can be scaled dependent on the problem. This now rescales the entries to have both continuous, positive, and negative weights. Other distributions for weights can be used, including discrete bi-valued or normal distribution, but uniform is generally a preferred distribution for its "continuity of values and boundedness [1]." This new adjacency matrix with defined weights will be referred to as $W_0$.

Next, as defined in [5], $W_0$ should be normalized by dividing each entry by $\rho(W_0)$, that is the spectral radius of the matrix where the maximum absolute eigen value, $|\lambda_{max}|$, is $\rho(W_0)$. This matrix is then scaled with the user specified spectral radius, $\rho(W)$; the spectral radius is one of the most important parameters in defining any ESN because it is related to the echo state property. Thus, reservoir weight matrix, $W$, is calculated as follows:

$$W = \frac{\rho}{|\lambda_{max}|} W_0 \qquad (3)$$

The reservoir activation functions to be examined are hyperbolic tangent, as it is the default reservoir activation for most ESN instantiations, and the normalized sinc function. The sinc function, as given in Equation 4, was chosen for this study as it showed promising improvement in a study done in [22]:

$$sinc(x) = \frac{\sin(\pi x)}{\pi x} \qquad (4)$$

Similarly, for the output activation function, hyperbolic tangent and identity activations will be studied, both of which are standard choices for ESN. For the experimentation, the following equations are employed:

$$x(t+1) = (1-\alpha)x(t) + \alpha f\left(W_{in}u(t+1) + Wx(t) + W_{fb}y(t)\right) + s_v v(t) \qquad (5)$$

$$y(t+1) = g(W_{out}[x(t+1)]) \qquad (6)$$

$$y(t+1) = g(W_{out}[1:u(t+1), x(t+1)]) \qquad (7)$$

$$y(t+1) = g(W_{out}[x(t+1), y(t)]) \qquad (8)$$

$$y(t+1) = g(W_{out}[1:u(t+1), x(t+1), y(t)]) \qquad (9)$$

The equation used in [13] to generate the NARMA-10 sequence is used in this study and is expressed by Equation 10, where $m$ is the random input sequence and $d$ is the NARMA-10 output sequence.

$$d(n+1) = 0.3d(n) + 0.05d(n)\left[\sum_{i=0}^{9} d(n-i)\right] + 1.5m(n-9) + 0.1 \qquad (10)$$

The Figure-8 sequence generated for this study is based on the work described in [2] and is expressed by Equations 11 and 12, where $ax\_x$ is the teacher sequence for the Figure-8 x-axis coordinates and $ax\_y$ is the teacher sequence for the Figure-8 y-axis coordinates. For both Equations $t$ is the current time step.

$$ax\_x(t+1) = \sin(2\pi t) \qquad (11)$$

$$ax\_y(t+1) = \cos(\pi t) \qquad (12)$$

The Mackey-Glass sequence generated for this study is based on the work described in [14] as well as [20] and is expressed by Equations 13 and 14, where Equation 14 specifically describes the starting sequence of $y$ for Equation 13 and was taken from [23].

$$\frac{\partial y}{\partial t} = (0.2y(t-\tau))/(1+y(t-\tau)^{10} - 0.1y(t)) \tag{13}$$

$$y = [0.9697, 0.9699, 0.9794, 1.0003, 1.0319, 1.0703, 1.1076, \tag{14}$$
$$1.1352, 1.1485, 1.1482, 1.1383, 1.1234, 1.1072,$$
$$1.0928, 1.0820, 1.0756, 1.0739, 1.0759]$$

Isolated Digits benchmark represents a time series classification problem, because of that the training differs slightly from the implementation given above. The classification training procedure used here is based on [1]. Instead of giving a prediction for every input sample, the ESN classifier takes an average of the current reservoir state over a given number of samples, and then returns a prediction; in this case, a group of samples is an isolated digit time series. Therefore, the notation for equations changes slightly for this problem as shown below:

$$y(t+1) = g(W_{out} \frac{1}{\tau} \Sigma[x(t+1)]) \tag{15}$$

$$y(t+1) = g(W_{out} \frac{1}{\tau} \Sigma[1: u(t+1), x(t+1)]) \tag{16}$$

$$y(t+1) = g(W_{out} \frac{1}{\tau} \Sigma[x(t+1), y(t)]) \tag{17}$$

$$y(t+1) = g(W_{out} \frac{1}{\tau} \Sigma[1: u(t+1), x(t+1), y(t)]) \tag{18}$$

$\tau$ in this case is the number of samples in a group. The revised notation is simply representative of the reservoir state, as well as inputs and outputs potentially, being averaged.

## Echo State Network Design Considerations

This section presents a general overview on overarching aspects of echo state network hyperparameters as well as various architectural changes that influence the performance. A variety of hyperparameters are employed for ESN based on architectural selections as well as how involved one wants to get with the optimization process. This discussion will encompass the minimum required hyperparameters for leaking ESN, with or without feedback connections.

The spectral radius, $p(\mathbf{W})$, is one of the most crucial selections for designing an ESN. It is the maximum absolute eigenvalue of the reservoir weight matrix. Defining the spectral radius scales the width of the distribution for the weights in the reservoir. In general, setting the spectral radius to a value between 0 and 1 will almost always guarantee the echo state property. There are very few cases where this does not work [24]. This is not to say that one cannot obtain the echo state property for values greater than unity; however, anything beyond unity will likely create instabilities in the network.

The leaking rate, $\alpha$, can be thought of as the time interval between two consecutive time steps; a representation from the continuous world realized discretely [1]. This term essentially gives one control over the memory of the reservoir as it allows neurons to retain parts of the previous states [5]. It can also be interpreted as the "speed of the reservoir update dynamics discretized in time [1]." The leaking rate can take on values in the interval (0,1]. A leaking rate equal to 1 removes the leaky integration and therefore becomes a simple ESN with no memory. Selecting a value for the leaking rate is dependent on the speed of the reservoir update dynamics of the input, $\mathbf{u}(t)$, and/or the output, $\mathbf{y}(t)$ [1]. As the value of this parameter is problem dependent, it must be set accordingly and yet its optimal value can be difficult to determine so the selection of the leaking rate then becomes a matter of trial and error.

The scaling of the input weight matrix, $\mathbf{W}_{in}$, is a key parameter for ESN optimization. It dictates the degree of nonlinearity to be represented by the reservoir state, $\mathbf{x}(t)$, as well as how much the current state is influenced by the current inputs versus the previous state, $\mathbf{x}(t-1)$ [1]. $\mathbf{W}_{in}$ is generated using some random distribution (usually uniform) with some density, $d_{in}$. The scaling for the inputs is not the same as the reservoir, which uses the spectral radius. For a random uniform distribution, $W_{in}$ can be defined on the interval $[-a, a]$, and for a normal distribution the standard deviation can be used to scale it [1]. For the scaling of $W_{in}$, one suggestion is scaling closer to 0 for linear tasks when using a hyperbolic tangent activation function, where the activations are essentially linear [1]. Likewise, larger scaler values will create non-linear behavior. Normally, there is a single parameter, $s_{in}$, defined for the scaling of $W_{in}$. To exert more control over how the inputs influence reservoir dynamics, one can create $s_{in}$ as a scaling vector so that any bias, as well as individual inputs, can be scaled separately.

Similarly, feedback weight matrix $W_{fb}$ can be discussed in the same context if one is implementing the optional feedback connections. Connecting feedback from the output into the reservoir is essentially adding more inputs into the reservoir; therefore, $W_{fb}$ needs to be scaled appropriately. The use of feedback should be restricted to solutions that absolutely must have it, like pattern generation problems, because using feedback often creates stability issues [2].

Reservoir size, $N$, is the parameter that defines the number of neurons within the reservoir. In general, larger values of $N$ require higher computational cost for the training effort but can provide higher levels of accuracy. Studies reported in the literature, [20],[5],[6],[11], indicate that a sparsely, and randomly connected reservoir weight matrix gives optimal performance. This means most of the connections in the reservoir, $W$, are non-existent. Creating a sparse reservoir generally does not influence the performance of the ESN to make a good prediction, but it does affect the computational efficiency for updating the reservoir [1]. A sparse reservoir enables fast reservoir updates. The density of $W$ is referred to as as $d_W$. Similarly, input and feedback weight matrices, $W_{in}$ and $W_{fb}$, can have their own densities, $d_{in}$ and $d_{fb}$, set as well. Unlike the reservoir though, these connections are typically dense [1].

Most commonly, uniform, discrete bi-valued, and Gaussian distributions are used to generate the weight matrices. Those who use uniform distributions for most of their work [1] state that it is preferred for its "continuity of values and boundedness." It is also argued that Gaussian distributions centered around zero typically have the same performance, dependent on parameter selection. Discrete bi-valued distributions do not give a rich signal space necessarily but have the potential to make the analysis of the reservoir dynamics easier to understand [1]. Overall, uniform distribution seems to be the standard for basic ESN setup, although some studies, like [25] and [22], do report using the Gaussian distribution. Experimentation with different weight distributions can lead to potentially better performing models.

## ESN Design for Simulation Study

In this section, the ESN will be configured for application to several different problem domains (pattern generation, chaotic systems, time series prediction, and time series classification) through explorations for various parameter values and architectural modifications. Accordingly, the following aspects of the ESN design will be studied:

- reservoir activation function ($f$),
- output activation function ($g$),
- optional connections to the output unit, and
- distribution of weights.

The Lazy Figure-8 and Mackey-Glass 17 tasks will require optional feedback connections, $\boldsymbol{W}_{fb}$. By default, all experiments will have a bias value of 1.0 injected into the reservoir neurons; in general, it is not a bad idea for any neural network to implement a bias term. It is suggested in [5] that adding a constant bias term will enable the ESN to train well for output mean values that deviate from zero. This choice was made to simplify the need to optimize an additional parameter. Parameters to be optimized are as follows:

- spectral radius ($\rho$),
- input scaling/feedback scaling ($s_{in}$ and $s_{fb}$, respectively),
- leaking rate ($\alpha$),
- the density of the reservoir/input/feedback weights ($d_W$, $d_{in}$, and $d_{fb}$, respectively), and
- the ridge regression regularization coefficient ($B$).

The size of the reservoir ($N$) will be held constant during the optimization process and will be adjusted later while examining the generated models. This study will also use ridge regression for training $\boldsymbol{W}_{out}$. The feedback matrix $\boldsymbol{W}_{fb}$ is used for the Lazy Figure-8 and Mackey-Glass 17 tasks, i.e. all weights in $\boldsymbol{W}_{fb}$ are zero for the rest of the benchmark problems, and a given experiment may opt out of either the input connected to the output unit or self-recurrent connections in the output unit. Note that the concatenation of vectors in Equations 7 and 9 includes a constant bias value of 1.0, where the notation $1: \boldsymbol{u}(t + 1)$ denotes that the bias value is appended to the input vector and is acting as an additional input into to the reservoir and output units. Lastly, it should be noted that Equation 5 appends a noise term for training and testing purposes but can simply be ignored and set to zero if no added noise is needed.

Random distributions to be examined for weight generation are uniform, discrete bi-valued, and Laplace centered at zero. Uniform and discrete bi-valued are common choices for weight distributions. Laplace centered at zero is similar to the commonly used Gaussian distribution but is not widely reported in literature.

**Setup for Simulation Study**

Python version 3.8.3 was used to implement an ESN class for full control over the architecture. Some of the expected parameters are number of input units, neurons in the reservoir, number of output units, spectral radius, etc. A few not so common parameters are leveraged to have full control over the architecture for experimentation. Those additional parameters specify the reservoir activation function, output activation function, the distribution used for weight generation, whether to add a bias value of one, whether to connect the input unit to the output unit, whether to connect the output unit into itself (self-recurrent), and whether this ESN will be used as a classifier. For more details on the implementation of this code, visit the GitHub repo.[1]

Since there will be a variety of different setups to test all these architectural changes, an optimization algorithm was implemented to expedite the search for viable combinations. Optuna version 2.6.0 [26] was chosen for the myriad of features it has for making hyperparameter optimization process manageable. It is comparable to HyperOpt, which is used in the ReservoirPy library [27], and is chosen here for its slightly more intuitive API. Optuna allows one to select from several different optimization algorithms, but defaults to the Tree-structured Parzen Estimator, which is a Bayesian optimization algorithm. This algorithm considers past evaluations to construct future hyperparameter trials and thus making it faster than the random grid search used in [27]. A study reports using a Gaussian process with Bayesian optimization in [28], which achieves optimal results in fewer iterations than other optimization methods.

The Optuna study object, which optimizes each of these ESN instantiations, is seeded with zero to help keep optimizations among different models more consistent and each study is comprised of a set of trials. The objective function passed to the study object defines the training process and returns the best root mean square error for the trial, or F1 score in the case of the Isolated Digit benchmark. Each trial runs the objective function once, but the objective function instantiates that specific ESN model ten times with different seeds and saves the best version. This is done because not every random generation of the reservoir will guarantee the echo state property.

---

[1] https://github.com/brooke-w/explore-esn

NumPy, a scientific computing library [29] version 1.19.2, is used for matrix and vector computations within the ESN class, as well as some data generation tasks and is mostly highly optimized C code. Lastly, all code has been run on the same computer using macOS Catalina version 10.15.7 with a 2.6GHz 6-Core Intel Core i7 processor, 16 GB 2667 MHz DDR4 memory, and an AMD Radeon Pro 5500M 4 GB Intel UHD Graphics 630 15136 MB graphics card. Experiments were created in the Spyder IDE version 3.3.6 and run with terminal. Results from these experiments were processed in JupyterLab version 3.0.16.

**Metrics for Performance Evaluation**

For all the regression benchmarks, namely NARMA-10, Lazy Figure-8, and Mackey-Glass 17, RMSE, MAE, and R2 score values were collected with the RMSE being used as the main evaluation metric. MAE and R2 scores were leveraged to confirm the performance of the model as indicated by the RMSE metric, i.e. give confidence in the RMSE score trends. R2 score was not chosen for the main evaluation criterion because it is not good at evaluating the stability of the model; this means for any model that deviates away from 'close to correct' for long periods of time, the score will drop considerably and likely become negative indicating a bad fit. This occurs with the Lazy Figure-8 and Mackey-Glass 17 benchmarks. R2 score values are evaluated in the range from 0.0 to 1.0, where the closer to 1.0 the better. However, the R2 score is a relative metric, a measure of variance in the model, making it a good candidate for comparing models trained on the same data.

One can gain a better sense of how well a model performs, even if it becomes unstable, with the RMSE metric because it stays within the range of 0.0 to infinity where the closer to 0 the better; if a model becomes unstable for a period, the score just increases versus becoming completely negative like the R2 score. It gives one a sense of increasing performance regardless of some bad predictions.

The MAE metric is the average of the absolute value of errors and gives one idea of how poorly the predictions of a model are in the sense of how far off they are from the actual value. The closer the MAE score to 0.0, the better the prediction. The RMSE metric penalizes errors more than the MAE, but otherwise they are very comparable and are good evaluation criteria. The RMSE score is used in our evaluations since it is the more frequently used metric, but MAE was also recorded to give confidence in the RMSE score.

The Isolated Digits benchmark is a classification problem so different metrics are used: F1 Score, Accuracy, and AUC. These metrics are all derived from the confusion matrix which represents true positives, false positives, true negatives, and false negatives. Here, the F1 score is used mainly in discussion of the results, but the selection of a classification scoring mechanism is highly dependent on what one wants to prioritize regarding true positives, false positives, true negatives, and false negatives.

Accuracy score prioritizes how many observations were classified correctly, returning a number 0.0 through 1.0, where the closer to 1.0 the better. However, this scoring mechanism does not consider false positives or false negatives so it can be misleading. The AUC score also returns a number between 0.0 and 1.0, the closer to 1.0 the better; it summarizes the results from the curve receiver operating characteristic curve which is plotted using the true positive rate (recall) and false positive rate at different thresholds. Lastly, F1 uses both recall and precision, making it more informative than accuracy since it takes both false positives and false negatives into account. F1 and AUC both incorporate true positives, true positives, false negatives, and false positives making them both good candidates for evaluation of the ESN models on the Isolated Digits data.

## Performance Evaluation Scenarios

### Time Series Prediction: NARMA-10

The setup described here is used to test a variety of ESN models for the NARMA-10 benchmark. Figure 2 shows a tree which displays all the model combinations of ESN architecture to be evaluated. These models will be identified in each study by a number sequence such as 1469; this corresponds to a path in the tree as they are labeled in the upper right corner, so this model would correspond to connections $u(t)$ and $x(t)$ to the output (Equation 7), tanh reservoir activation function ($f$), a identity output function ($g$), and a random uniform distribution used for the reservoir generation. Each model was optimized using Optuna [26].

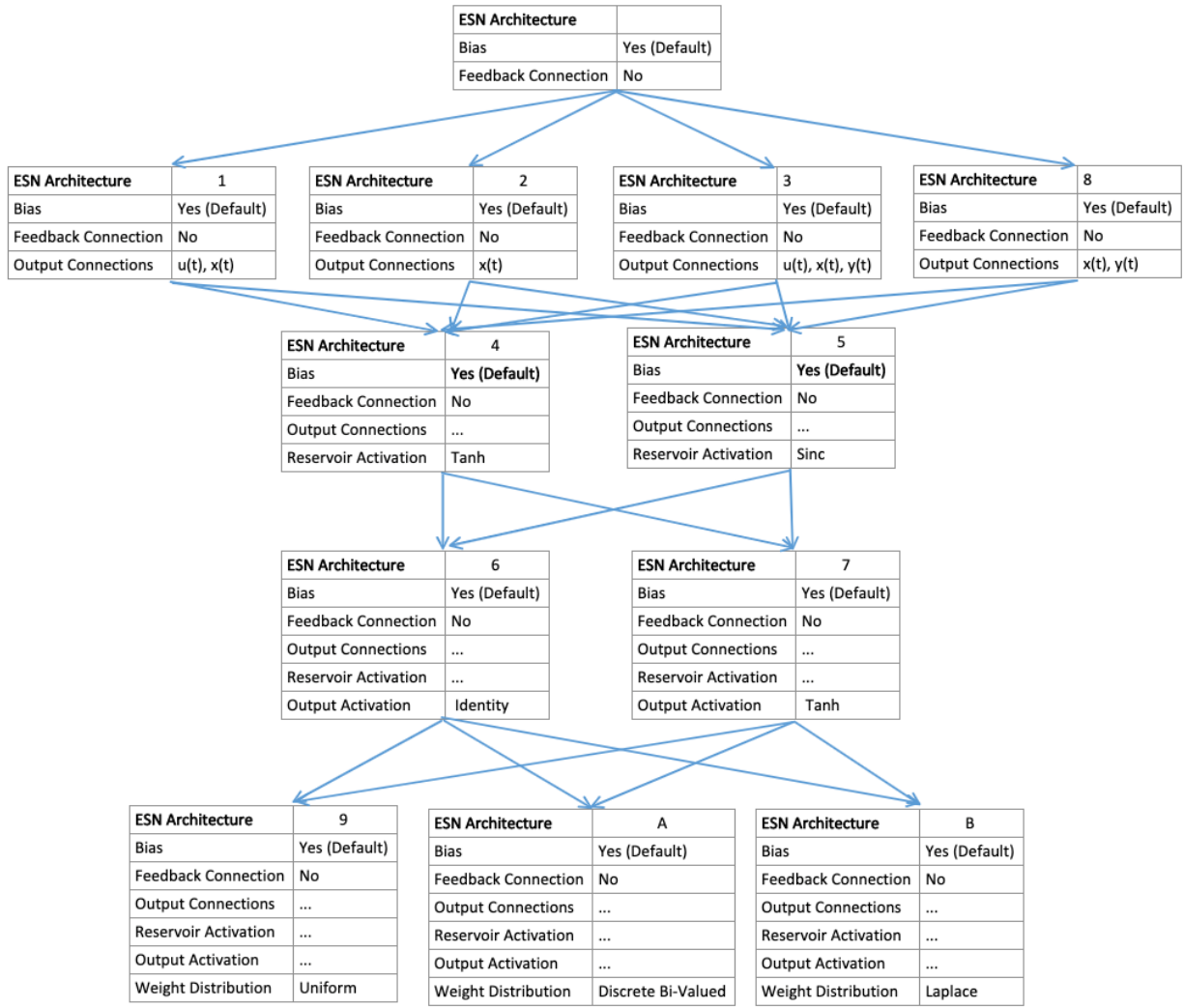

**Figure 2:** A tree representing all the different architectural and parameter-value pair changes to be made to the ESN. A total of 48 possible models are represented by this tree. Each study will be labeled according to the number in the upper right corner of each node to identify its attributes. Example: study_1469 would correspond to connections $u(t)$ and $x(t)$ to the output (Equation 7), tanh reservoir activation function ($f$), a identity output function ($g$), and a random uniform distribution used for the reservoir generation.

The NARMA-10 sequence used in the study is expressed by Equation 10. The NumPy library [29] was used to generate a random uniform sequence from [0.0, 0.5) and was seeded with zero for a consistent generation of NARMA-10 sequence. Figure 3 shows the first 100 timesteps of output for this NARMA-10 sequence generation approach.

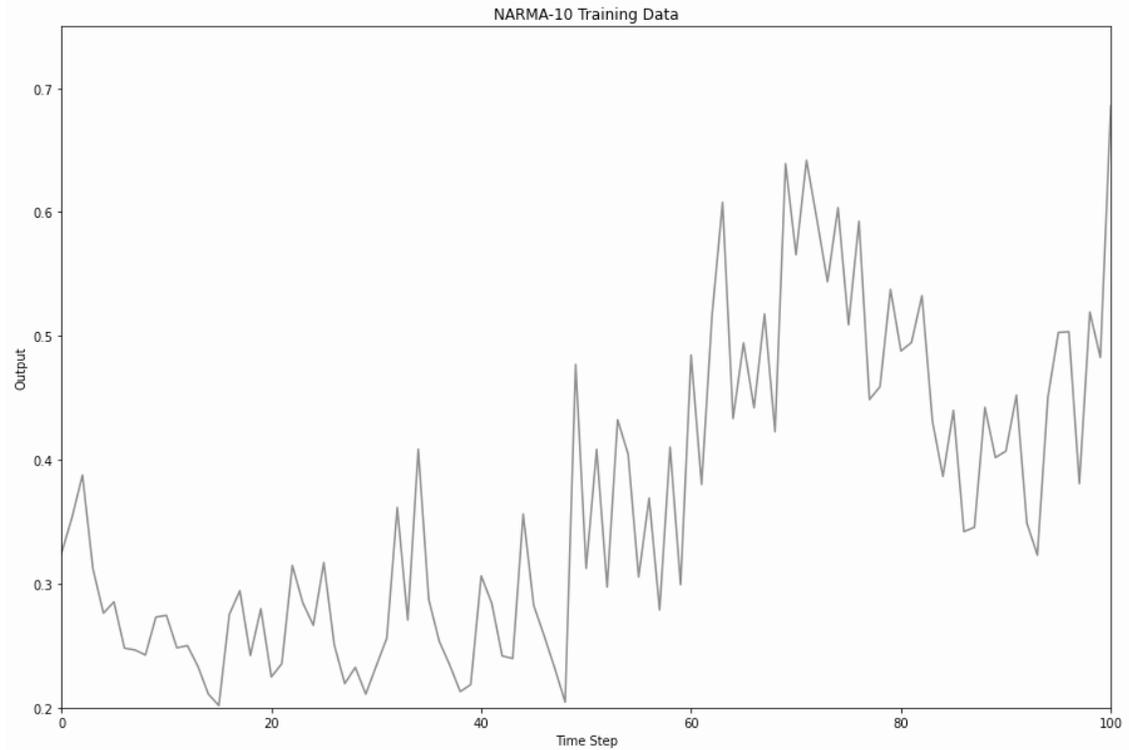

**Figure 3:** The NARMA-10 data created with Equation 10 with input $m$ as the random input sequence and $d$ as the output sequence. For $m$, the NumPy library[29] was used to generate a random uniform sequence from [0.0, 0.5) and was seeded with zero for a consistent generation of NARMA-10.

The training methodology used in these simulation experiments was based on [13]. The NARMA-10 dataset is made up of 3400 samples, 1200 of which are in the training set and the other 2200 are in the test set. A washout of 200 samples was used for both training and testing. Random uniform noise of the range [-0.0001, 0.0001) was injected into the reservoir during training (i.e. $v$ was generated using a random uniform distribution from [-1,1) and $s_v$ was set to 0.0001 to scale). Note, this study uses ridge regression unlike [13] so, although noise is used here, it could have been ignored since the regularization coefficient helps alleviate noise and overfitting sensitivities. The value of $N$, the number of neurons in the reservoir, was set to 100 during the Optuna optimization, as in [13]; it is sufficient for generating accurate predictions, which is what is needed for preemptive model results, and the size can be altered later to test how it influences accuracy.

For the NARMA-10 benchmark, 150 trials per study were run to find an optimized set of parameter values; this value was decided on through preliminarily testing of these trials to ensure reasonably good parameter values would be found. Parameters to be optimized are the spectral radius ($\rho$), leaking rate ($\alpha$), density of the reservoir ($d_W$), density of the input weights ($d_{in}$), scaling of the input weights ($s_{in}$), and the regularization coefficient ($\beta$). For this benchmark problem, feedback is not necessary so the scaling ($s_{fb}$) and density ($d_{fb}$) of the feedback weights can be set to zero. Other miscellaneous parameters include number of inputs ($K$) and outputs ($L$), which were both set to one, as well as parameters for defining the architecture for each model described by Figure 2.

**Pattern Generation: Lazy Figure-8**

Figure 4 shows a tree which displays all the ESN model combinations subjected to performance evaluation and can be read the same way as the tree described for the NARMA-10 benchmark testing. This problem necessitates the use of feedback and the only input in this case is the bias into the reservoir and, dependent on the model, the bias into the readout layer. The Figure-8 sequence is generated by Equations 11 and 12. Figure 5 shows the first 200 timesteps of output, which is also the full Figure-8 sequence.

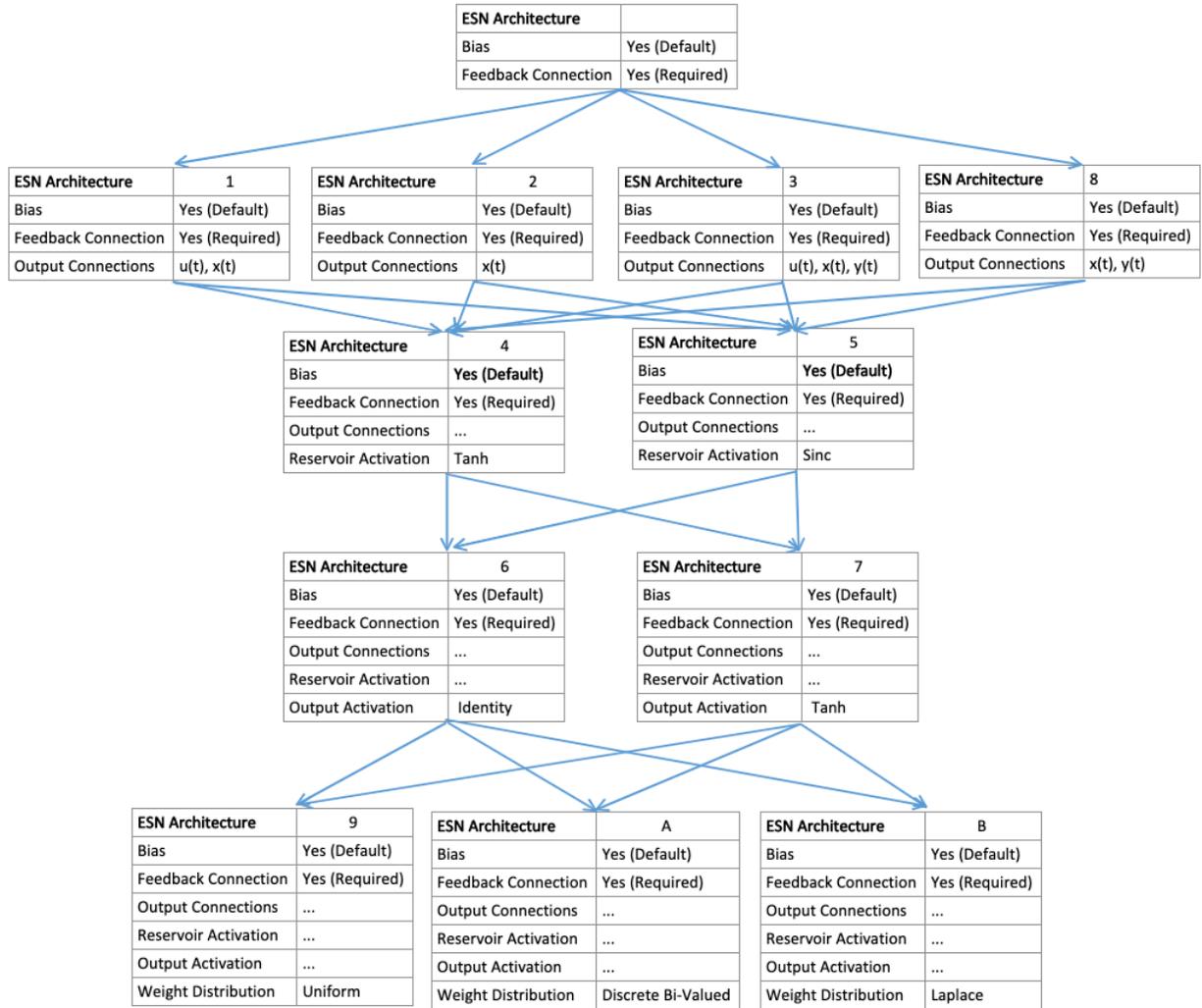

**Figure 4:** A tree representing all the different architectural changes to be made to the ESN. A total of 48 possible models are represented by this tree. Each study will be labeled according to the number in the upper right corner of each node to identify its attributes. Example: study_1469 would correspond to connections $u(t)$ and $x(t)$ to the output (Equation 7), tanh reservoir activation function ($f$), a identity output function ($g$), and a random uniform distribution used for the reservoir generation.

The training used in these simulation experiments was based on [2]. The Lazy Figure-8 dataset is made up of 23,000 samples, 3,000 of which are in the training set and the other 20,000 are in the test set. A washout of 1000 was used for both training and testing to ensure all initial transients were removed from the reservoir; this value could have been less, but since data in this case is easy to generate and the goal is to get a stable, continuous pattern generation, having a larger washout will guarantee those transients are removed. The testing data set is much larger than the training because, once again, the goal is to produce a stable model, so it needs to be tested over a relatively long duration. Random uniform noise was injected into the reservoir during prediction, not during training (i.e., $v$ was generated using a random uniform distribution from [-1,1) and $s_v$ was set to 0.01 to scale). Noise is injected during the prediction period to test the overall stability of the pattern generating ESN model.

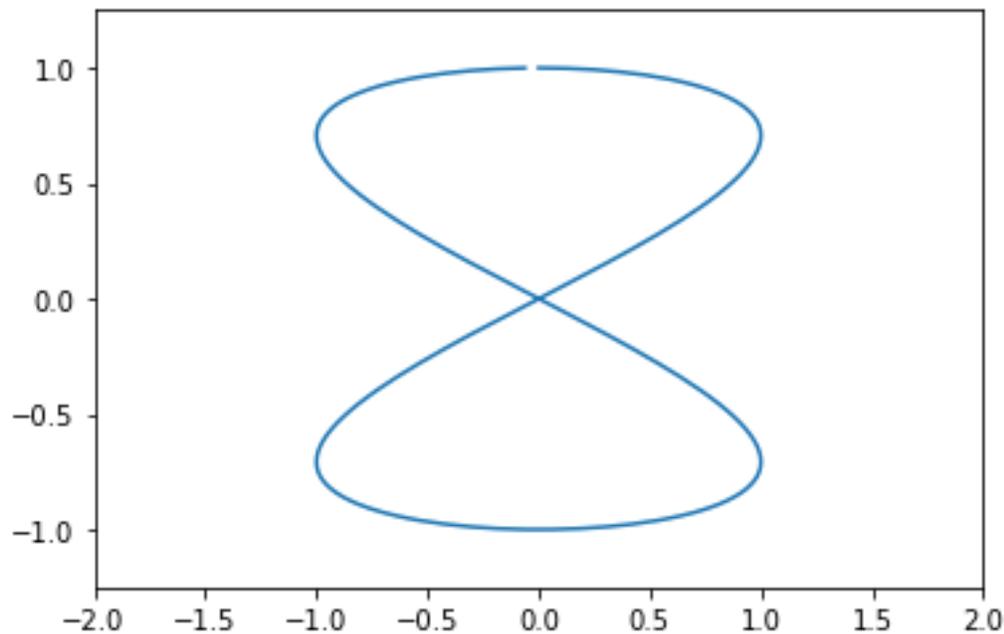

**Figure 5:** The Lazy Figure-8 generation created with Equations 11 and 12 for the first 200 timesteps, $t$.

For the Lazy Figure-8 benchmark, 150 trials per study were run to find a good set of parameter values; any less and a perfectly good model may never find satisfactory parameter values. For the optimization, the size of the reservoir, $N$, was kept constant at $N = 20$. Parameters to be optimized are the spectral radius ($\rho$), leaking rate ($\alpha$), density of the reservoir

($d_W$), density of the input weights ($d_{in}$), density of the feedback weights ($d_{fb}$), scaling of the input weights ($s_{in}$), scaling of the feedback weights ($s_{fb}$), and the regularization coefficient ($\beta$). Other miscellaneous parameters set includes number of inputs ($K$) and outputs ($L$), which were zero and one respectively, as well as parameters for defining the architecture for each model in Figure 4.

**Chaotic System: Mackey-Glass 17**

This problem requires the use of feedback and the only input in this case is the bias into the reservoir and, dependent on the model, the bias into the readout layer. The Mackey-Glass sequence is generated by Equations 13 and 14. A total of 4,100 samples were generated as shown in Figure 6 where the first 100 were dropped.

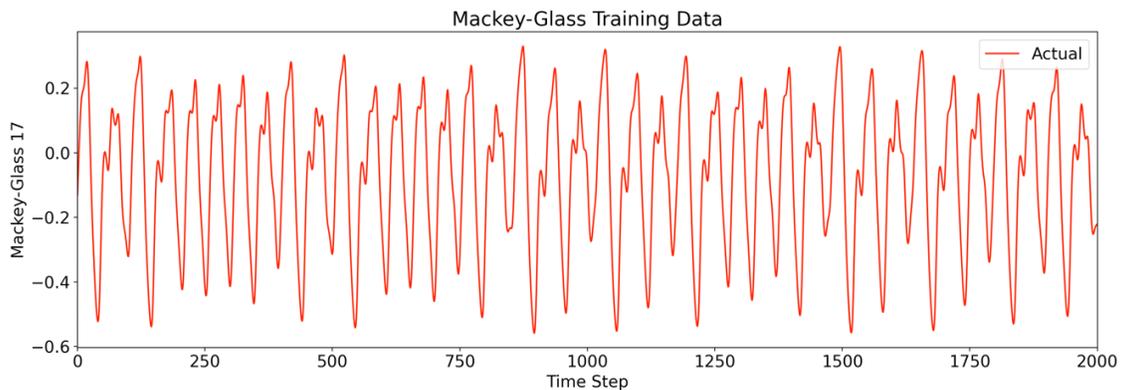

**Figure 6:** The Mackey-Glass training sequence created with Equation 13.

The training methodology used in these simulation experiments was based on [20] with some time saving alterations. For the 4,100 samples in the Mackey-Glass 17 dataset, 2000 samples are in the training set and the other 2,000 are in the test set, where the test set is the continuation of the training set sequence. A washout of 100, which is a high enough value to ignore transients in this case, was used for only training, as the goal here is to have the ESN continue to generate that Mackey-Glass sequence using feedback connections. The ESN implementation saves the last state of the reservoir unless it is explicitly set or washed out during the prediction step, so there is no need for washout in this case. No noise was injected for training and prediction; the ridge regression coefficient should assist with instabilities introduced by feedback connections instead.

For the Mackey-Glass 17 benchmark, 150 trials per study were run to find a satisfactory set of parameter values. Parameters to be optimized are the spectral radius ($\rho$), leaking rate ($\alpha$), density of the reservoir ($d_W$), density of the input weights ($d_{in}$), density of the feedback weights ($d_{fb}$), scaling of the input weights ($s_{in}$), scaling of the feedback weights ($s_{fb}$), and the regularization coefficient ($\beta$). Other miscellaneous parameters to set include number of inputs ($K$) and outputs ($L$), which were set to zero and one, respectively, as well as parameters for defining the architecture for each model.

**Time Series Classification: Isolated Digits**

Each ESN architectural model combination is individually optimized using Optuna [26]. The data for this benchmark is from [30], an open-source repository on GitHub. This dataset emulates the TI46 isolated digit dataset which is used in [12] and [9]. It has a total of 6 speakers pronouncing English digits 0 through 9, where each speaker pronounces each digit 50 different times for a total of 3,000 recordings overall. Constraints on this dataset were put in place to decrease the overall runtime for experiments. The ESN models were only trained and tested with digits 0 through 4. For the optimization experiments, 60% of the overall data was used to train ESN and 20% was dedicated to testing, for a total of 31,360 samples, or 600 recordings, in the training set and 15,550 samples, or 300 recordings, in the test set; with the recordings being randomly sorted, this led to 174 digit 0 recordings, 189 digit 1 recordings, 177 digit 2 records, 186 digit 3 recordings, and 174 digit 4 recordings for the training set, and 53 digit 0 recordings, 61 digit 1 recordings, 61 digit 2 records, 52 digit 3 recordings, and 73 digit 4 recordings for the test set. To reduce runtime for data collection on larger reservoir sizes, all generated graphs were trained with 40% of the overall data with 10% used for testing; with the recordings being randomly sorted, this leads to 120 digit 0 recordings, 129 digit 1 recordings, 121 digit 2 records, 119 digit 3 recordings, and 111 digit 4 recordings for the training set and 29 digit 0 recordings, 29 digit 1 recordings, 25 digit 2 recordings, 33 digit 3 recordings, and 34 digit 4 recordings for the test set.

The data was preprocessed using the Lyon passive ear model from [31], which is a Python port of the one in reference [32]. This method is mentioned in [12], [9], and [16], however the number of frequency channels differs among these studies. Here, 85 frequency channels are used which is like the one in reference [12]. A single isolated digit pronunciation has a myriad of samples that represent the change in the 85 frequency channels over time, so a group identifier is given to each sample and the samples within the group are kept in chronological order. The target information for this classification problem is set up as follows: for each given sample, there are five possible targets which should all be 0 except for one which contains 1, or 0.99999, identifying the isolated digit being pronounced. The representation of 1 was 0.99999, a number very close to 1.0, to test the tanh output activation function which cannot be used for output that is undefined for arctanh.

Because this is a classification problem, the training differs slightly from the implementation, therefore, the notation for Equations 15, 16, 17, and 18 is used for this problem: A visual representation of these isolated digit time series groups is given in Figure 7 as a spectrogram, which shows how frequency varies with time.

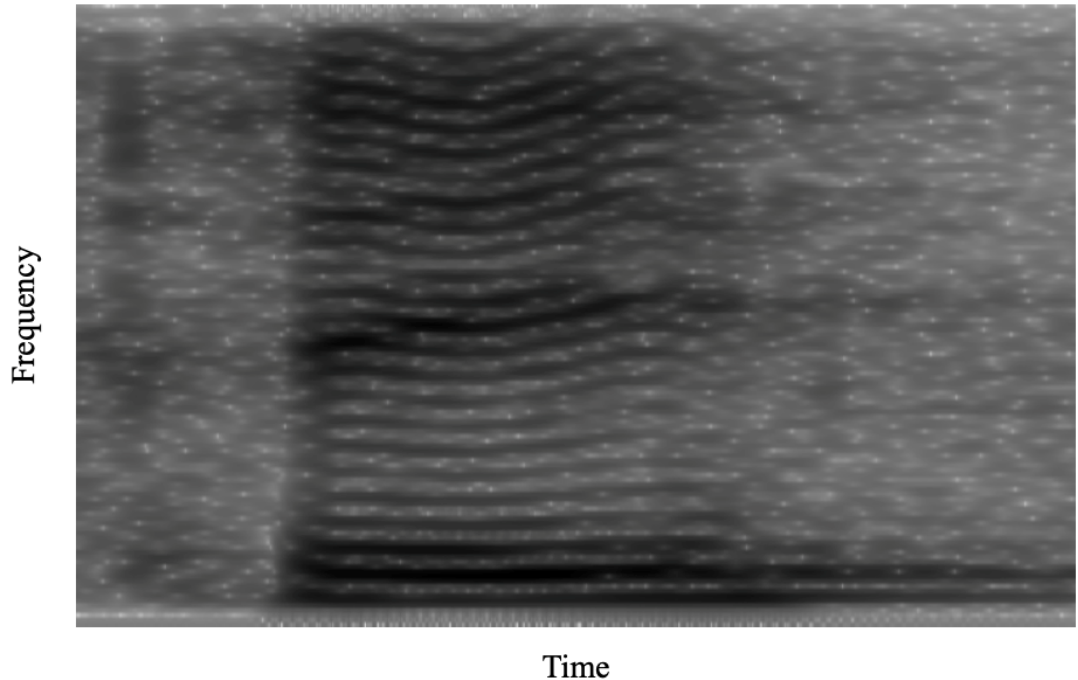

**Figure 7:** A spectrogram of speaker George pronouncing 'three'. The darker lines are representative of frequencies present.

For the optimization, the reservoir size was held constant at 50 neurons. In [16], 200 neurons are used in the reservoir for this type of problem; however, this slows down optimization of each study. The expectation is that for a good set of parameters an increased reservoir size should increase the prediction quality of the ESN, so this setup is fine for finding a good set of parameters. No noise is injected into the reservoir, which is typically done through simulated noise to give a more general fit, as these audio samples already have trace amounts of noise in them.

For the Isolated Digits benchmark, 50 trials per study were run to find an optimized set of parameter values. A smaller number of trials were run for this benchmark as it did not seem to need the greater number of trials to find a good set of parameter values and greatly reduced runtime for the optimization process. Parameters to be optimized are the spectral radius ($\rho$), leaking rate ($\alpha$), density of the reservoir ($d_W$), density of the input weights ($d_{in}$), scaling of the input weights ($s_{in}$), and the regularization coefficient ($\beta$). For this benchmark problem, feedback is not necessary so the scaling ($s_{fb}$) and density ($d_{fb}$) of the feedback weights can be set to zero. Other miscellaneous parameters include number of inputs ($K$) and outputs ($L$), which were 85 and 5, respectively, as well as parameters for defining the architecture for each model.

## Simulation Results

Evaluation of the overall performance of ESN models for all four benchmarks problems was done. Optimization for a good set of parameter values for all models except a few was successful and those models performed reasonably well for a large enough reservoir. Each ESN model with its optimized parameters had the number of neurons in the reservoir increase several times, starting at 50, and with 100 neuron increments. For each increase in the neuron count, the root mean square error was averaged over 15 different seeded instantiations (with outliers removed) for each model. From literature ([1],[3],[22]), it was expected that the increase of neurons for any well performing model with a small reservoir will improve the performance, i.e. decrease the root mean square error. Figure 8 presents agreement with literature findings, showing trends toward increasing accuracy for models that showed good performance for smaller reservoir sizes.

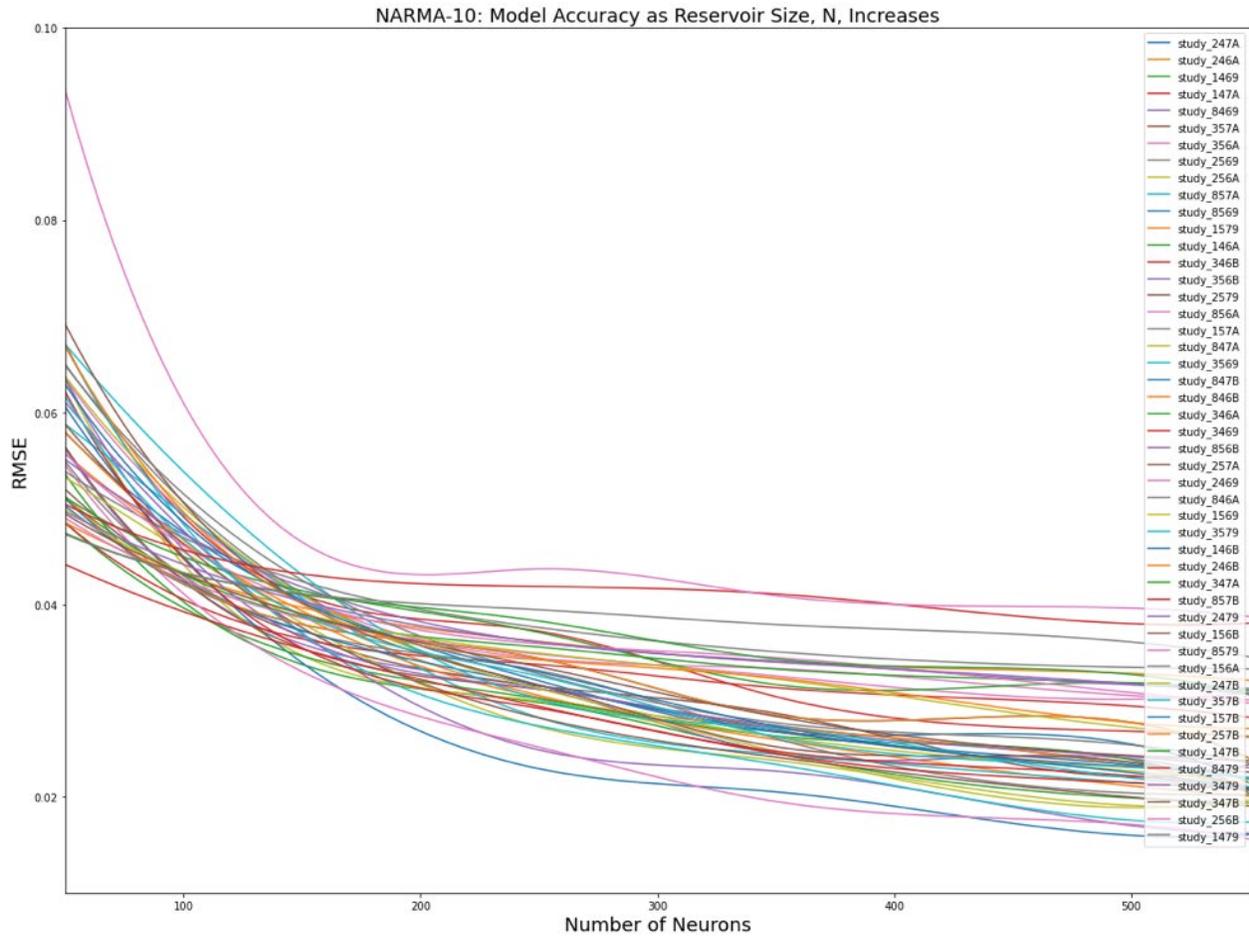

**Figure 8:** The performance of optimized NARMA-10 models for increasing reservoir size, $N$, measured using RMSE, where each model was run for reservoir sizes 50, 150, 250, 350, 450, etc., for 15 seeded instantiations which were averaged with outliers removed.

One can gather from Figure 9 that training time is increasing polynomially as the number of neurons in the reservoir increases. Plotted alongside the training runtimes from each study are curves for $O(N^c)$ and $O(Nlog(N))$. One can see that the polynomial function, $O(N^c)$, is a good fit, therefore, the training time complexity for an ESN model can be categorized as $O(N^c)$, where $N$ corresponds to the number of neurons in the reservoir and $c$ is some constant. In this case, $c$ is approximately 2. In Figure 10, one does not see the same trend; the prediction step has a constant runtime for each input sequence that increases linearly with $N$; there is spiking that occurs around $N = 1000$ which is due to hardware limitations. To keep the results consistent, all experiments and results were generated on the same computer, but it was confirmed on the high-performance computing cluster that the linear trend continues. One can categorize the prediction time complexity of an ESN as $O(N)$. The time complexity difference between training and prediction can be attributed to the generation of the reservoir weights, $W$; the larger the reservoir becomes, the more computationally expensive the generation due to the complexity of finding the eigenvalues for scaling the matrix, which has polynomial time complexity. Since the reservoir is only generated once, this process does not need to be repeated for a specific model.

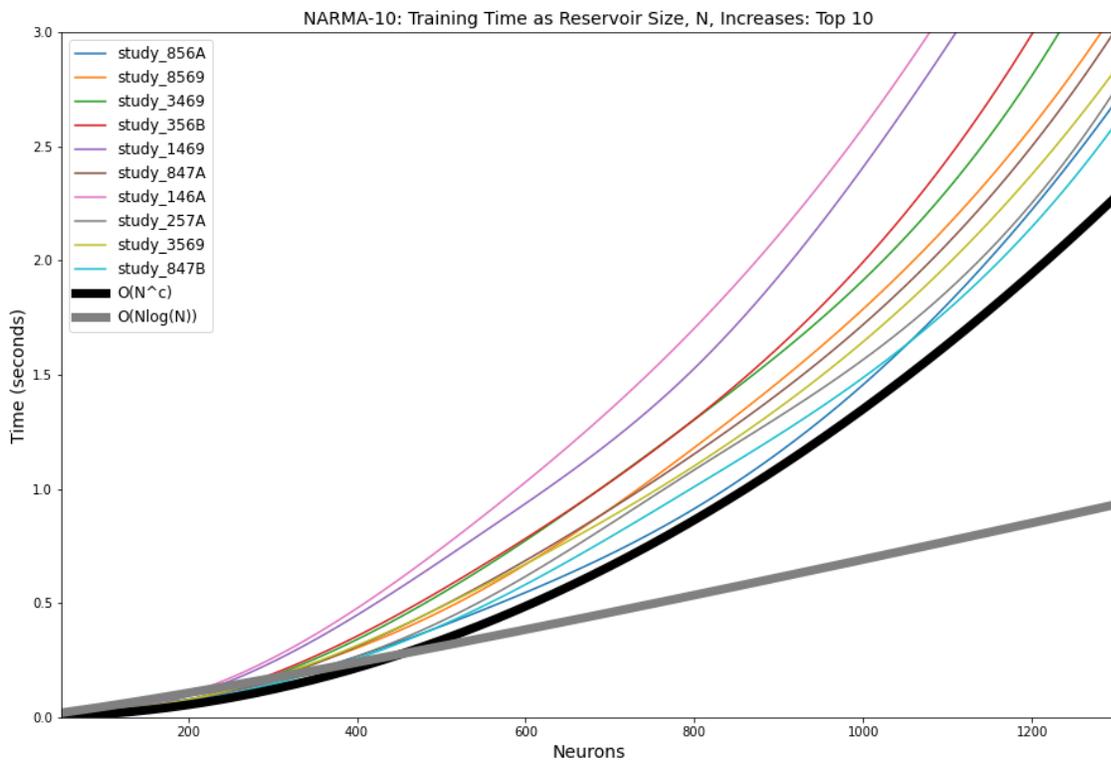

**Figure 9:** Top ten NARMA-10 models selected by the lowest average of all the RMSE scores for each reservoir size per model. This figure shows the training runtime of each optimized model for increasing reservoir size, $N$, where each model was run for reservoir sizes 50, 150, 250, 350, 450, and 550, for 15 seeded instantiations which were averaged with outliers removed. The black line shows $O(N^c)$ and the gray line shows $O(Nlog(N))$. Here, $c$ is approximated as 2.

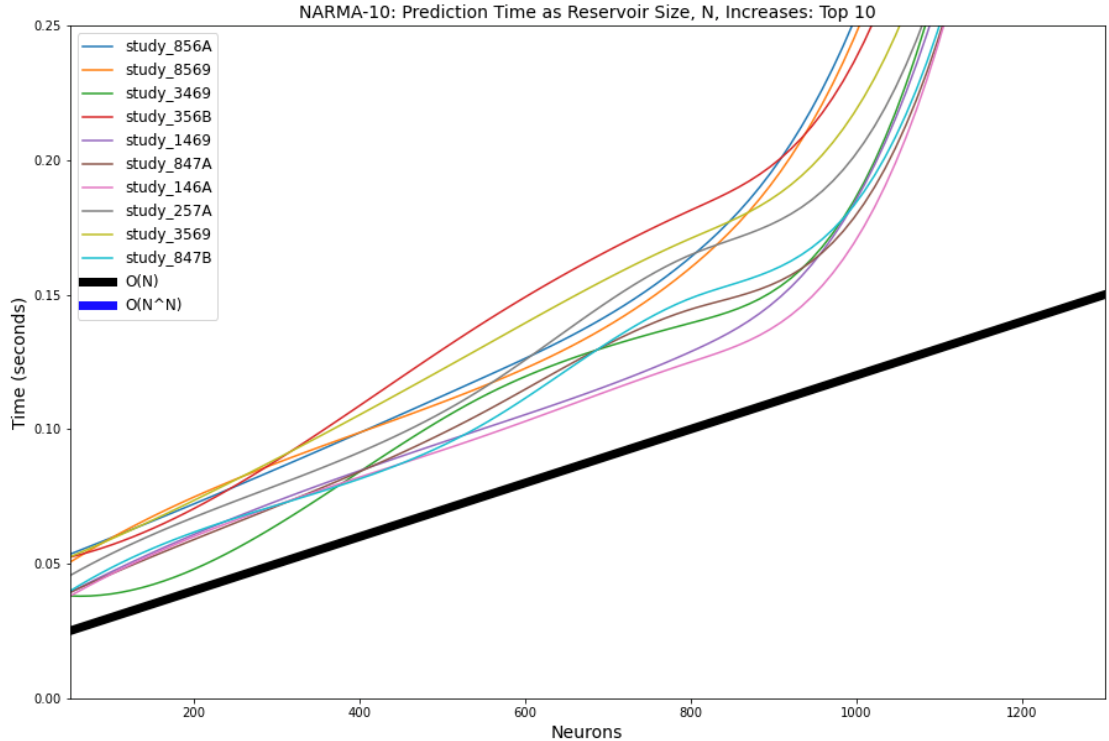

**Figure 10:** Top ten NARMA-10 models selected by the lowest average of all the RMSE scores for each reservoir size per model. This figure shows the prediction runtime of each optimized model for increasing reservoir size, $N$, where each model was run for reservoir sizes 50, 150, 250, 350, 450, and 550, for 15 seeded instantiations which were averaged with outliers removed. The black line shows $O(N)$, or linear runtime. The spike occurring around 1000 neurons is due to hardware constraints; it was confirmed the trend does continue linearly on the high performance computing cluster.

Figure 11 shows the performance of 29 optimized models for increasing reservoir size, where each model was run for reservoir sizes 50, 150, ..., to 550. For each reservoir size, a model was run for 15 differently seeded instantiations where RMSE for each of the runs was averaged together with outliers removed. Unlike NARMA-10, the Lazy Figure-8 models do not all perform similarly. In fact, most of the instantiations are chaotic and unstable. This is better seen in Figure 12, where the R2 score was used to evaluate the performance using the same process described for Figure 11. There are two models in particular, study_2569 and study_1579, that seem to just not perform well at all, and their scores hover around or just below 0.0. Most of these models expose their inability to keep stability with negative R2 scores. A few studies, like study_246B and study_256B, with a large enough reservoir seem to improve enough to give stable generations while others never improve, as their performance does not change as the reservoir grows. Specifically, study_356B shows exceptionally chaotic behavior that a smooth trend line could not be fit to it.

      The time complexity trends for the Lazy Figure-8 benchmark are the same as those shown for NARMA-10. Figure 13 suggest a polynomial time complexity, $O(N^c)$, curve fit to the training time as the reservoir size increases, where $N$ is the reservoir size and $c$ is some constant; in this case, $c$ can be approximated as 2. Figure 14 shows the prediction time complexity of the Lazy Figure-8 benchmark, which trends linearly with $N$ just as NARMA-10 does; there is a spike that occurs around $N = 1000$ which is due to hardware limitations. To keep the results consistent, all experiments and results were generated on the same computer, but it was confirmed on the high-performance cluster that the linear trend continues. One can categorize the prediction time complexity of an ESN as $O(N)$. At a quick glance, it is difficult to determine what is really influencing any small-time discrepancies between these models.

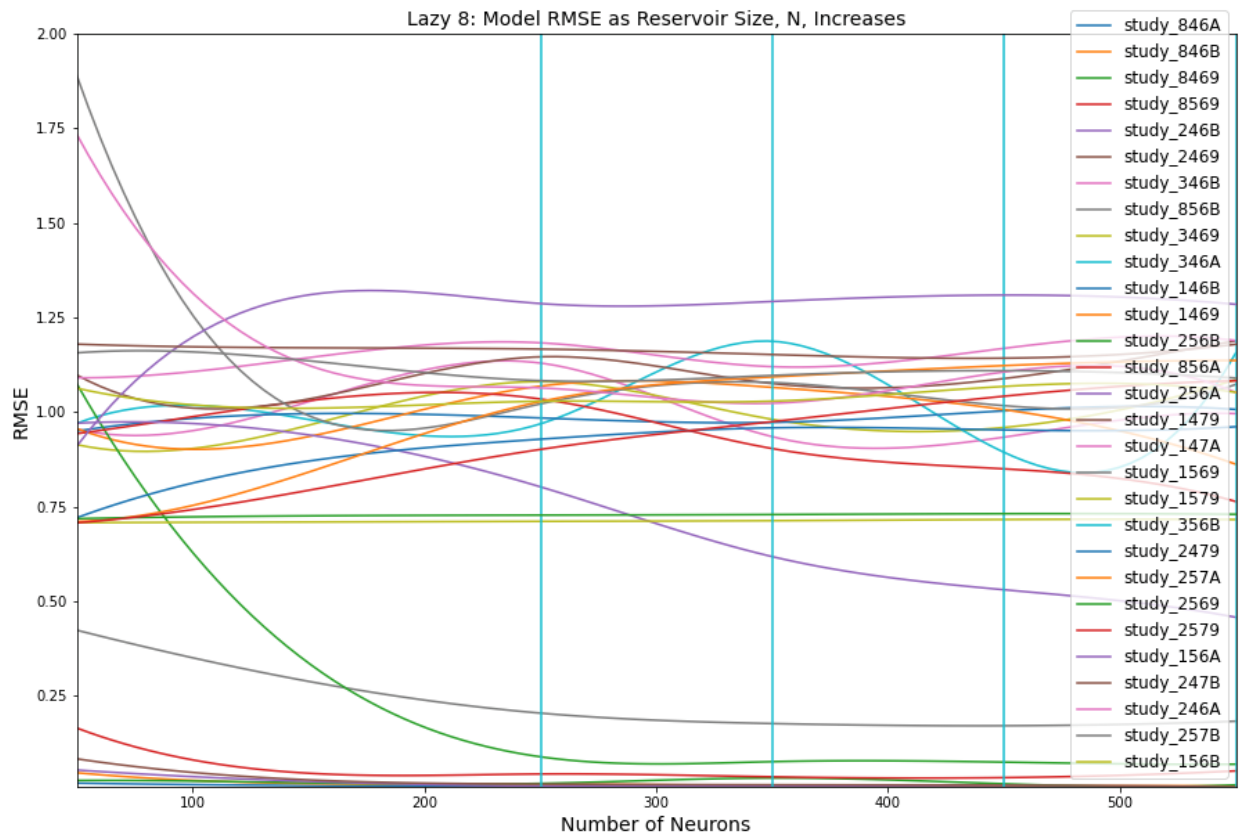

**Figure 11:** The performance of optimized Lazy Figure-8 models for increasing reservoir size, $N$, measured using RMSE, where each model was run for reservoir sizes 50, 150, 250, 350, 450, and 550, for 15 seeded instantiations which were averaged with outliers removed. Here, only 29 of the 48 available models were plotted.

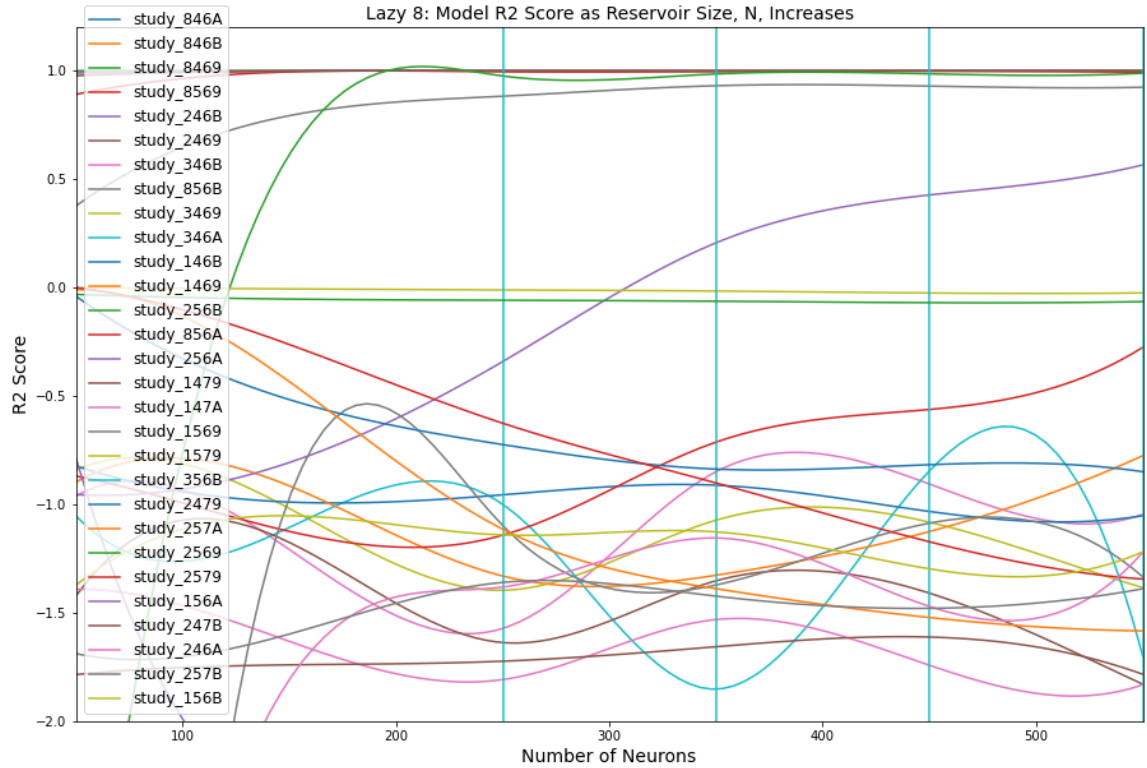

**Figure 12:** The performance of each optimized Lazy Figure-8 model for increasing reservoir size, $N$, measured using R2 score, where each model was run for reservoir sizes 50, 150, 250, 350, 450, and 550, for 15 seeded instantiations which were averaged with outliers removed.

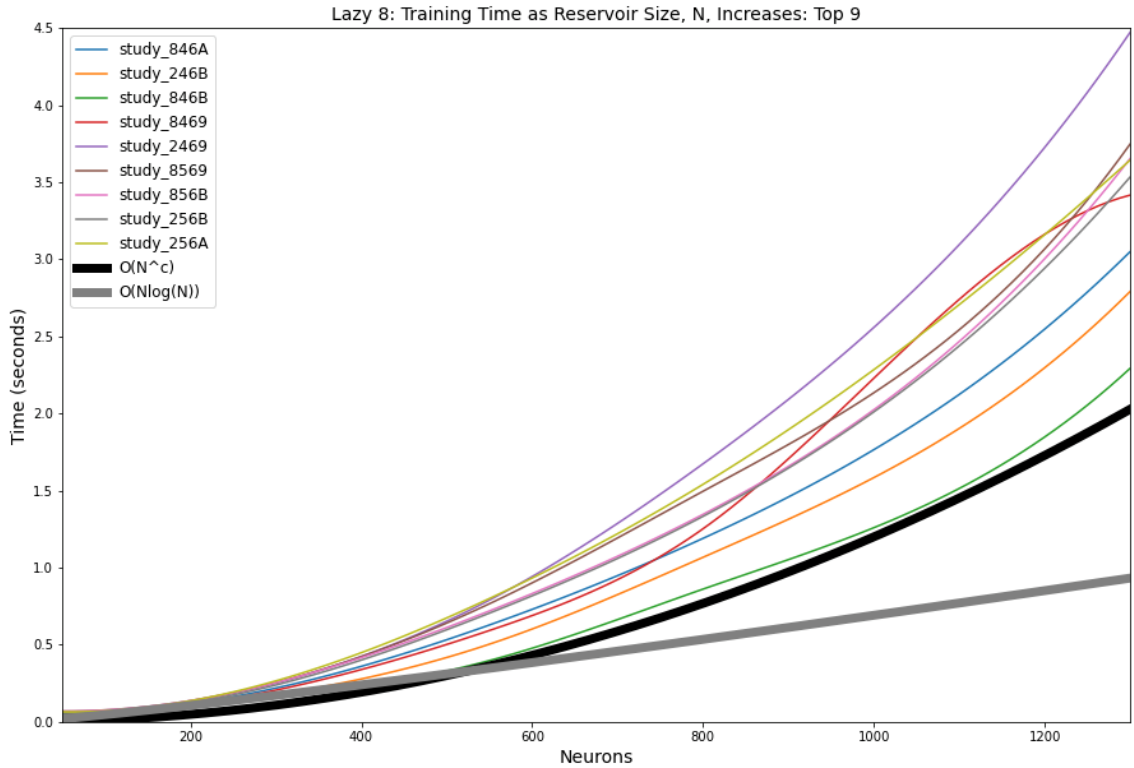

**Figure 13:** This figure shows the training runtime of the top nine optimized models for increasing reservoir size, $N$, where each model was run for reservoir sizes 50, 150, 250, ..., and 1350 for 15 seeded instantiations which were averaged with outliers removed. The black line shows $O(N^c)$ and the gray line shows $O(Nlog(N))$. The top nine Lazy Figure-8 models shown in this figure were selected by the lowest average of all the RMSE scores for each reservoir size per model.

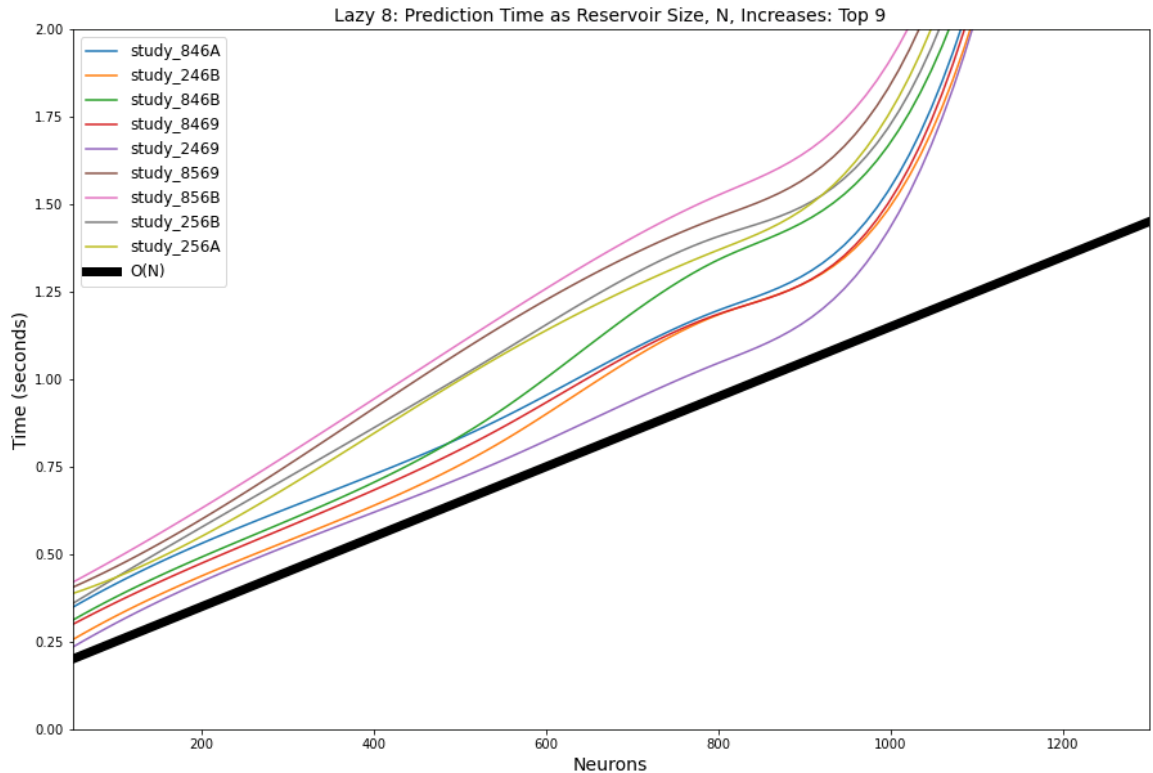

**Figure 14:** This figure shows the prediction runtime of each optimized model for increasing reservoir size, $N$, where each model was run for reservoir sizes 50, 150, 250, ..., and 1350, for 15 seeded instantiations which were averaged with outliers removed. The black line shows $O(N)$, or linear runtime. The top nine Lazy Figure-8 models shown in this figure were selected by the lowest average of all the RMSE scores for each reservoir size per model. The spike that occurs around 1000 neurons is due to hardware constraints; the trend does continue linearly when run on the high-performance computing cluster.

Figure 15 shows the RMSE values for each optimized study as the reservoir size grows. For any good set of parameter values, the RMSE value settles to a more consistent value. However, for some models like study_856B and study_3469, it seems that the curves do not settle toward any particular value. This can be explained by extremely large RMSE values for small reservoir sizes; for a large enough reservoir, those models have acceptable RMSE value, but the library being used to smooth the plotted line between sample points has a hard time creating a nice smooth line with the initially large RMSE value from reservoir sizes of 50 and 150. All models except study_346A found a good set of parameter values and performed reasonably for a large enough reservoir.

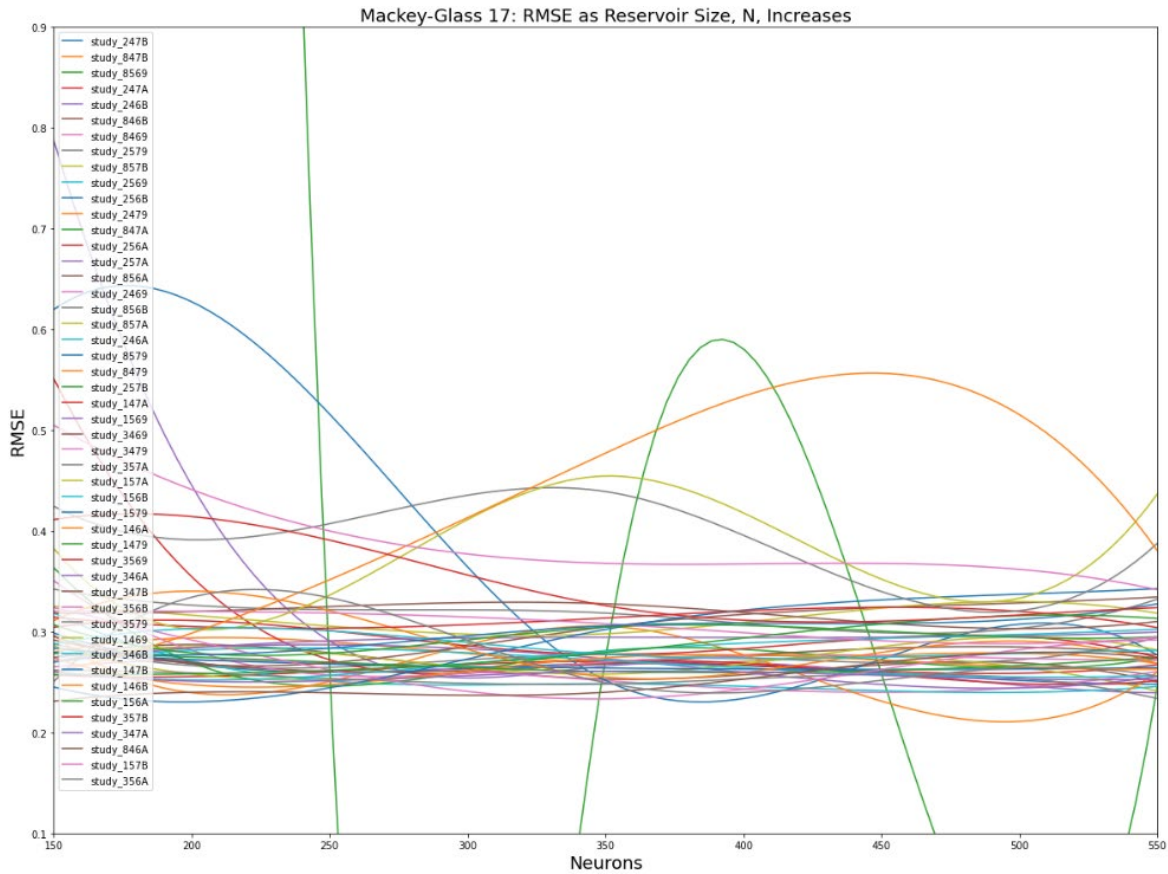

**Figure 15:** The performance of optimized Mackey-Glass 17 models for increasing reservoir size, $N$, measured using RMSE, where each model was run for reservoir sizes 50, 150, 250, 350, 450, and 550 neurons, for 15 seeded instantiations which were averaged with outliers removed.

Next, the time complexity of ESN is evaluated with respect to the Mackey-Glass 17 benchmark. Figure 16 reveals a polynomial time complexity curve, $O(N^c)$, when evaluating the training time as the reservoir size increases. In this case, $N$ is the size of the reservoir, or number of neurons, and $c$ is some constant which is estimated to be 2. Figure 17 shows the prediction time complexity as the reservoir size increases. One can see the prediction time complexity increases linearly with the reservoir size as $O(N)$.

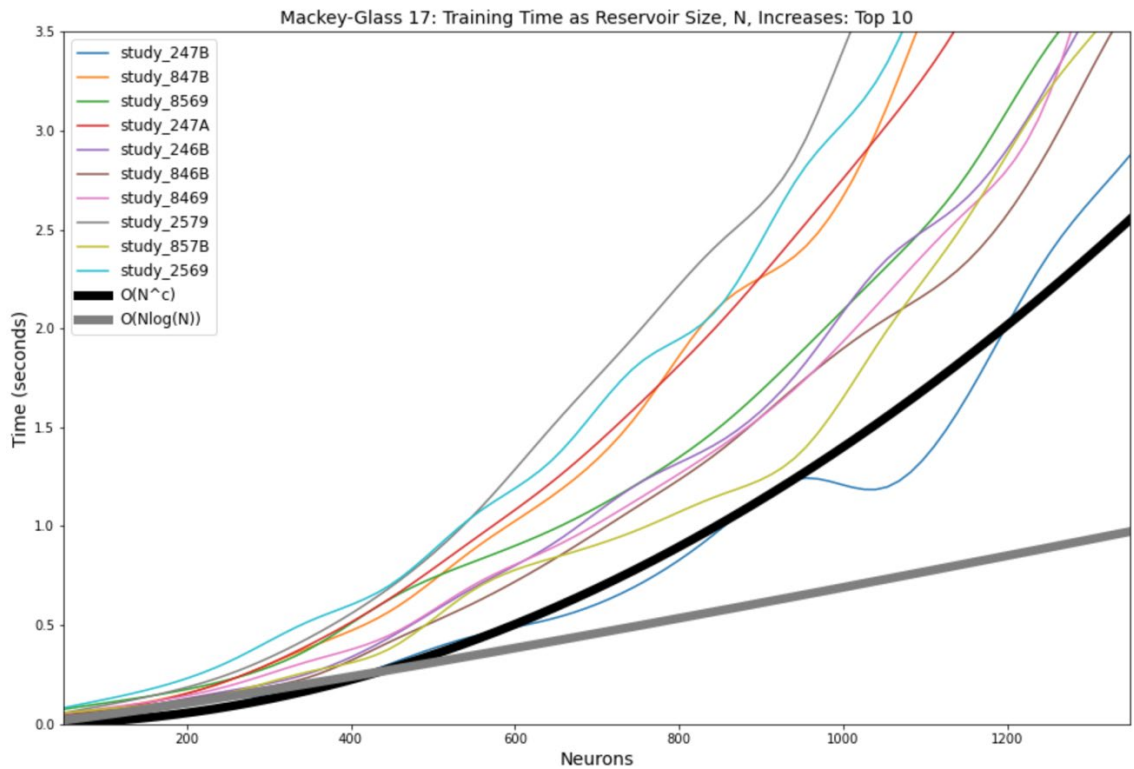

**Figure 16:** This figure shows the training runtime of each optimized model for increasing reservoir size, $N$, where each model was run for reservoir sizes 50, 150, 250..., and 1350 neurons for 15 seeded instantiations which were averaged with outliers removed. The black line shows $O(N^c)$ and the gray line shows $O(Nlog(N))$. The top ten Mackey-Glass 17 models shown in this figure were selected by the lowest average of all the RMSE scores for each reservoir size per model.

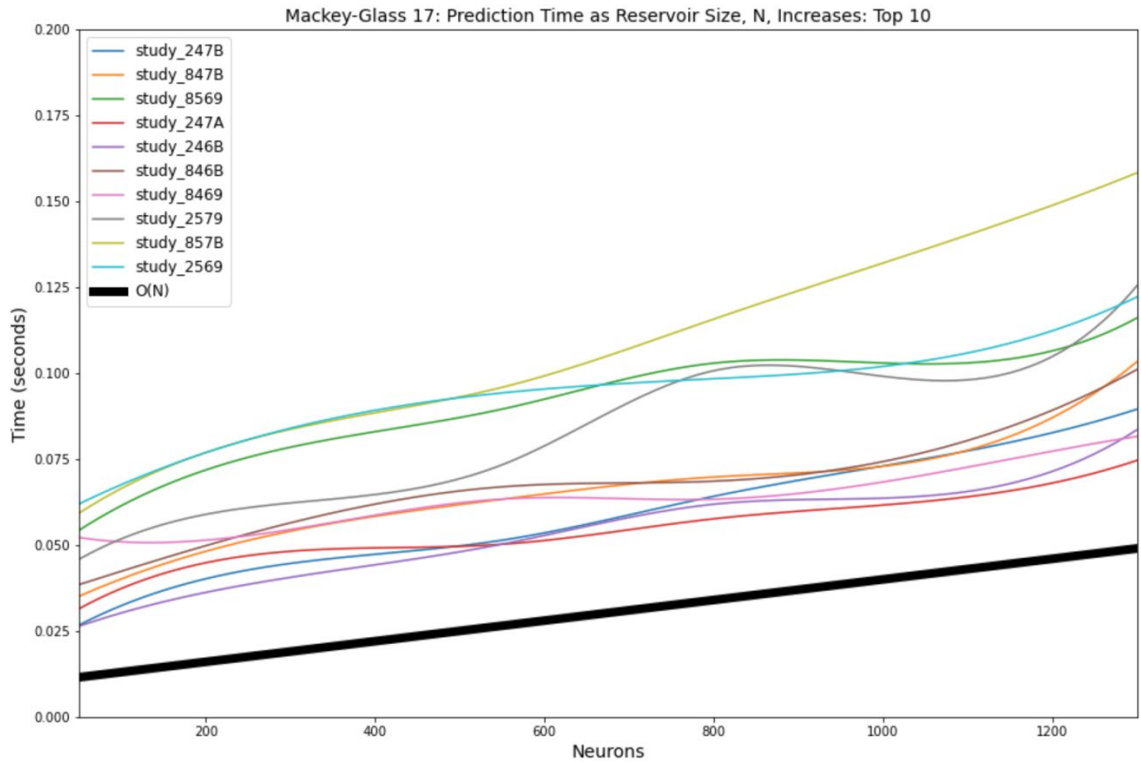

**Figure 17:** This figure shows the prediction runtime of each optimized model for increasing reservoir size, $N$, where each model was run for reservoir sizes 50, 150, 250, ..., and 1350 neurons, for 15 seeded instantiations which were averaged with outliers removed. The black line shows $O(N)$, or linear runtime. The top ten Mackey-Glass 17 models shown in this figure were selected by the lowest average of all the RMSE scores for each reservoir size per model.

Figure 11 displays F1 metric values for the top ten performing models for the isolated digits problem. The top ten models are defined as those that showed an overall increase in the F1 score for increasing reservoir size or the model had one of the highest F1 scores because not all of them performed consistently as the reservoir size increased. In fact, most of the models chosen for the top 10 do not show improvement in prediction performance for increasing reservoir size.

This time series classification problem does not have a consistent trend of increasing prediction accuracy, in terms of F1 score, as the reservoir size increases except for a few studies. Figure 19 shows the AUC score for these models for increasing reservoir size up to 1050 neurons. Once again, the same trends shown for the F1 score are present with the AUC score as some showing increasing AUC values for larger reservoirs. Obviously, it was possible to find a good set of parameters through the optimization experiments for a reservoir size of 50, which was smaller than what was reported in literature. The number of trials in each experiment may have needed to be extended beyond 50 to guarantee a good set of parameter values such that the F1 score or AUC score continues to improve as the reservoir grows, as the performance is somewhat consistently higher for these parameter sets and for smaller reservoir sizes.

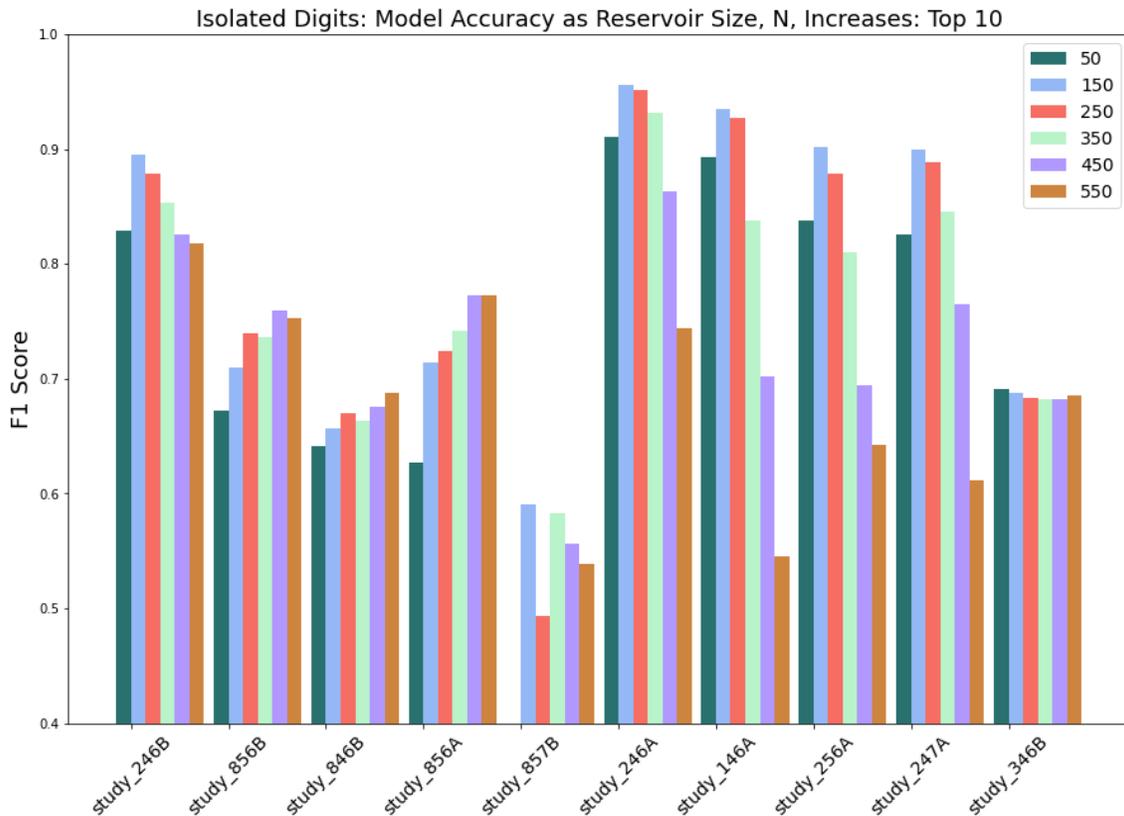

**Figure 18:** Top ten Isolated Digit models that were selected for an overall increase in the F1 score for increasing reservoir size or the model had one of the highest F1 scores.

The performance of each optimized model for increasing reservoir size, $N$, where each model was run for reservoir sizes 50, 150, 250, 350, 450, and 550, for 15 seeded instantiations.

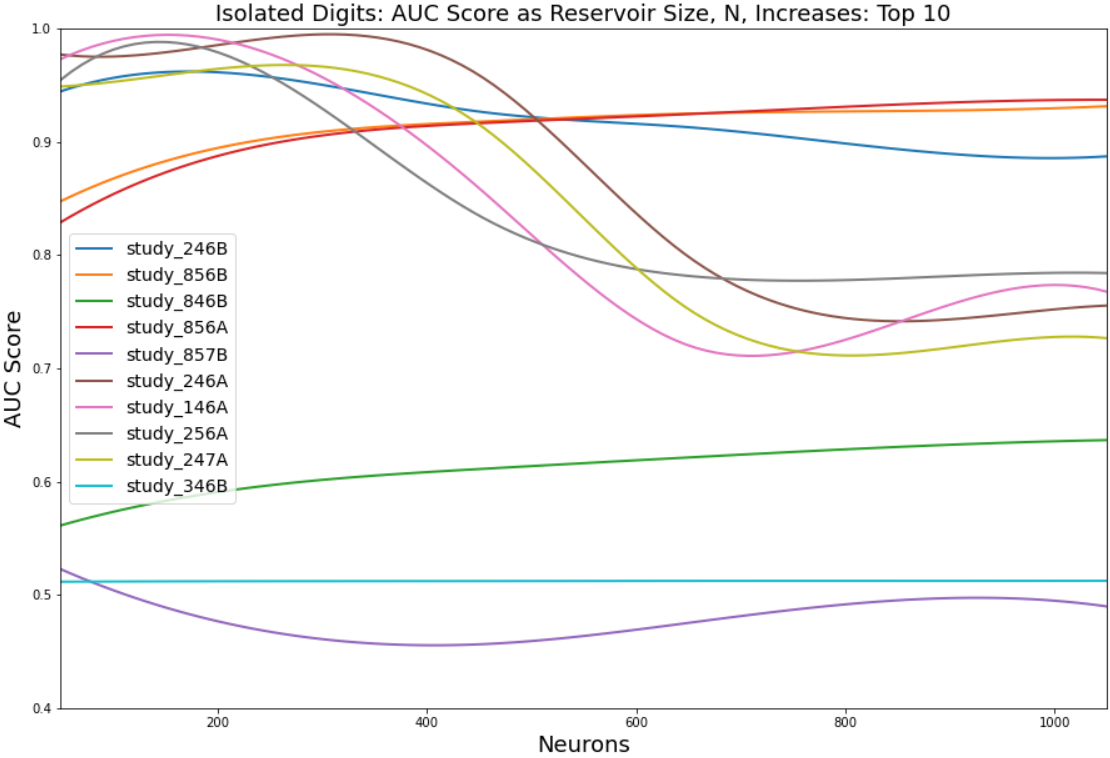

**Figure 19:** Top ten Isolated Digit models' AUC scores for increasing reservoir size, that were selected for an overall increase in the F1 score for increasing reservoir size or the model had one of the highest F1 scores. The performance of each optimized model for increasing reservoir size, $N$, where each model was run for reservoir sizes of 50, 150... and 1050 neurons, for 15 seeded instantiations.

Figure 20 shows the training time complexity for the Isolated Digit benchmark for the top 10 models. It seems that as the reservoir size increases, the time complexity increases polynomially or $O(N^c)$, where $c$ is approximately 2. However, unlike previous benchmarks, the polynomial curve does not seem to be quite as close of a fit. In fact, considering some of the hardware constraints previously encountered at reservoir sizes larger than 900 neurons, it may be said that a line may look to be a better fit. This linear complexity can be explained by the copious amounts of data fed to this model versus the previous benchmarks. The only operation that is polynomial in nature is the generation of the reservoir, but here the training on the data greatly outpaces that time it takes to generate the reservoir, so the linear complexity is easily observable. Therefore, for a large enough reservoir, one would once again see this curve clearly trending as $O(N^2)$. Similarly, Figure 21 shows the prediction runtime complexity for the Isolated Digit benchmark for the top ten models. As the reservoir size increases, prediction runtime increases linearly, or $O(N)$.

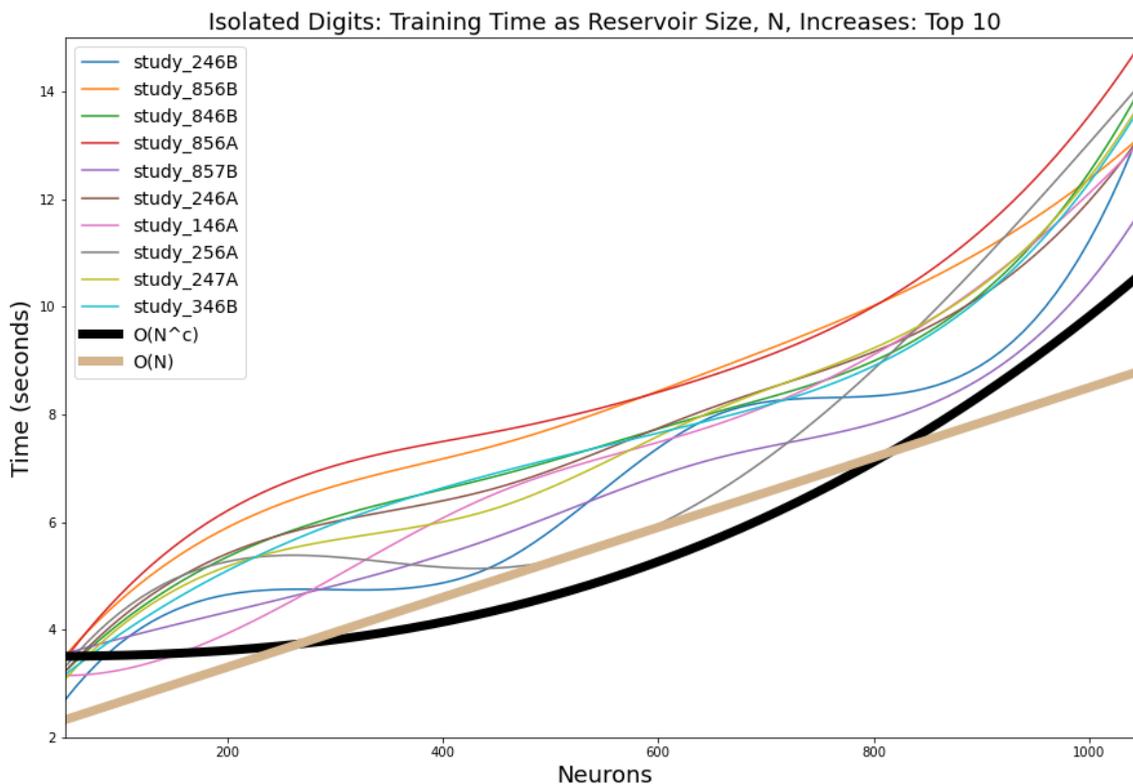

**Figure 20**: Top ten Isolated Digit models that were selected for an overall increase in the F1 score for increasing reservoir size or the model had one of the highest F1 scores. This figure shows the training runtime of each optimized model for increasing reservoir size, $N$, where each model was run for reservoir sizes 50, 250, 450, ..., 1050 neurons, for 15 seeded instantiations which were averaged with outliers removed. The black line shows $O(N^c)$ where $c$ is approximately 2. The tan line shows $O(N)$.

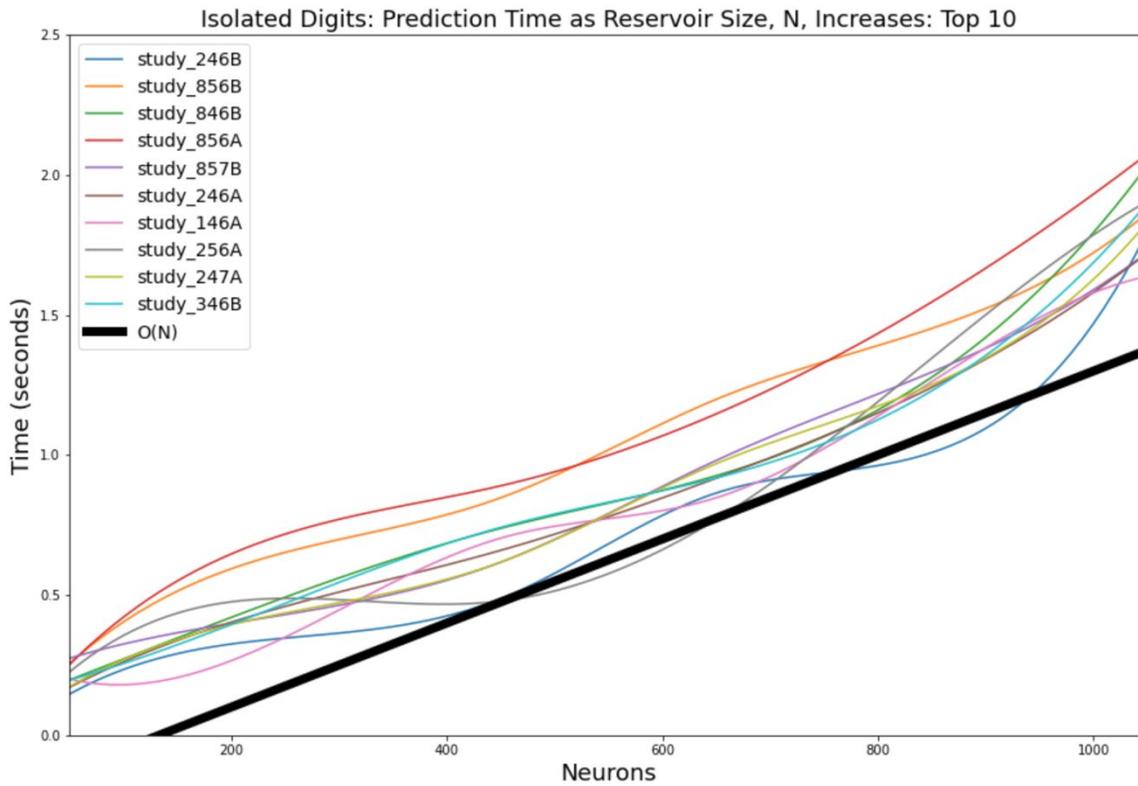

**Figure 21:** Top ten Isolated Digit models that were selected for an overall increase in the F1 score for increasing reservoir size or the model had one of the highest F1 scores. This figure shows the prediction runtime of each optimized model for increasing reservoir size, $N$, where each model was run for reservoir sizes 50, 250, 450, ..., 1050 neurons for 15 seeded instantiations which were averaged with outliers removed. The black line shows $O(N)$, or linear runtime.

**Observations**

Across all these four studies, there were several prevailing trends. For all these studies, polynomial training runtime complexity was observed for increasing reservoir size, $O(N^2)$, as well as linear prediction runtime complexity, $O(N)$, with the exception of the Isolated Digit benchmark. To get a more precise approximation to the time complexity curve, one must look at the generation of the reservoir; this is the only place where the training of the ESN differs from generating predictions in terms of time complexity. The NumPy library[29] is used to generate the eigenvalues of the randomly generated reservoir and, it seems to be using a variation of the QR algorithm to compute the eigenvalues, which has an $O(n^3)$ time complexity. One can also incorporate the fact that the training phase has some identical functionality to the prediction phase within it, adding the linear component $O(N)$. Therefore, it would be more accurate to conclude that the training time complexity of the ESN presented here, in relation to the reservoir size, for the worst case is $O(N^3 + N)$.

On the same note, runtime complexity was shown to only be influenced by one parameter consistently among all benchmarks, except the Isolated Digits benchmark, and that was the reservoir density. These benchmark problems also showed that the reservoir density, $d_W$, did not need to be set anywhere close to 1.0, a fully connected reservoir, to achieve good results. However, the closer the density got to 1.0, the training runtime noticeably increased for a large enough reservoir. Increasing the density of the reservoir did not increase the prediction quality for any of the benchmark problems. Keeping the reservoir sparse, around 0.15 and 0.20, was sufficient, which is in line with what is reported in reference [1].

## Guidelines for ESN Design

The empirical evidence due to these benchmark experiments provides the information needed to understand the complexities of various echo state network parameters, their values, as well as changes to the ESN architecture. Specific rules of thumb, or guidelines, can be provided for each problem to assist with the creation of ESN models for similar problem domains, as well as the background knowledge needed to understand where they are derived from. These guidelines are highlighted as follows:

- There is no need to make a reservoir extremely dense. For a good set of parameters, a sparse reservoir performs just as well. From these observations, approximately 0.15 for all studies performed just as well as larger values.
- Increasing the reservoir size, $N$, is an easy way to improve performance, but there is a time trade-off and, at some point, the improvement to be had by increasing the reservoir size may not be worth it. Plotting the RMSE as $N$ increases, as well as the runtime, helps establish what size the reservoir should be.
- Keeping the $N$ value small while exploring parameter and architectural options is preferable. Many parameter sets worked just as well for larger reservoir sizes. Keeping $N$ small during the setup phase will keep runtimes shorter.
- $N$ does have to be large enough for the ESN to work 'well enough' to find a good set of parameters, so if an ESN is not functioning as expected while keeping $N$ small, try increasing the reservoir slightly.
- A denser input weight matrix for input only problems (i.e. no feedback) seems to overall assist in the predictive capabilities of an ESN and does not greatly influence the runtime. It is recommended to set $d_{in} \geq 0.90$, and, if it is to be optimized, it is low priority.

- o  A denser input weight matrix for problems that used feedback was observed to not always enhance performance when set to be dense. With inputs and feedback in use, there is more optimization to be done when it comes to balancing the input to feedback ratio. One may just opt to keep $d_{in} \geq 0.90$ and attempt to adjust this ratio through the input scaler term, but the density should be looked at eventually to optimize as well.

- No feedback problem required a fully dense feedback weight matrix. There is a balance that must be obtained between the inputs and feedback coming into the reservoir. This may be best achieved by setting set $d_{fb} \geq 0.90$ and manipulating the feedback scaler accordingly but optimizing the feedback density may have to be revisited still.
- For any regression problem that has some dependency on previous outputs, self-recurrent connections should be added.
- An identity readout activation is sufficient given a good set of parameters.
- Hyperbolic tangent or the sinc functions both suffice for good reservoir activation function choices given the right set of parameters. The echo state property influences the ability of these functions to make use of the information in the reservoir as well as the reservoir parameters. Given this information, one may take different approaches to selecting a reservoir activation:
  - Find a good set of parameters for the default tanh reservoir activation function, then swap out the activation function with sinc, or another activation function of choice. Check the performance of each activation function as the reservoir size increases.

- Create an ESN model for using each reservoir activation function and optimize the spectral radius and leaking rate for each model. Compare these models for increasing reservoir size.

- For all the regression problems (NARMA-10, Lazy Figure-8, and Mackey-Glass 17), for the right set of parameters, all distributions could achieve similar RMSE values.

- Only the classification problem, Isolated Digit, showed noticeable preference for discrete bi-valued and Laplace distributions so one may opt to start with one of those distribution options for other time series classification problems.

- Higher spectral radius values close to 1.0 were observed to enhance performance in problems that required an extensive history of the inputs.

- Similarly, problems that relied on recent history saw spectral radius values closer to 0.0. These would be problems that require predictions on short duration time series data.

- For the regularization coefficient, the value is going to be dependent on the problem, but it was observed for problems that use feedback to keep generating predictions that higher values tend to help alleviate error from propagating throughout the reservoir. It was also observed to assist in generalizing the fit to noise prone data at higher values.

- Connecting a bias to the reservoir overall seems to assist the model in making better predictions. One may want to experiment with different bias values or assigning a scaler to the bias for more control, which was not done in this study.

- Larger scaler values for input and feedback increase non-linearity in the reservoir so it is recommended to start at small values and increase the values accordingly until no improvement can be observed.
- Lastly, the leaking rate enables memory and rates close to 0.0 indicate heavy reliance on the previous states of the reservoir, while values closer to 1.0 indicate reliance on the current inputs and reservoir state. Setting this value is going to be dependent on the data one is working with. Below are some tips for setting this value based on these observations:
    - Pattern generation or similar problems will need a leaking rate closer to 0.0 as the pattern generation is highly dependent on where the model is in the pattern generation process.
    - For short time series that are highly reliant on previous states, a higher leaking rate close to 1.0, implying some retention of previous samples but placing importance on the most current sample. This prevents initial transients, or potentially information from previous time series groups if one did not set the reservoir state back to zero, to be retained in the reservoir therefore influencing the prediction.
    - Memory is not necessarily needed for all problems to get a good prediction. In this case, setting the leak rate to 1.0 and removing it may be sufficient and simplifies obtaining the echo state property.

**Conclusions**

In this work, we have presented exploratory application of Echo state network to representative problems in four domains namely NARMA time series prediction task, Lazy Figure-8 pattern generation task, Mackey-Glass 17 chaotic system, and the Isolated Digit time series classification task. The objective was to identify, define or formulate guidelines, heuristics, or value ranges for hyperparameters of the network as well as most appropriate architectural topologies to help mitigate the relatively steep learning curve associated with the application of this type of neural network architecture to problems in similar domains. The empirical study based on simulations resulted in findings along these lines as presented. It is therefore reasonable to expect that these empirically derived guidelines will be useful towards cutting down on the need for exploration in the hyperparameter and architectural topology spaces of the Echo state network for its application to a specific problem.

## Data Availability

Data sharing is not applicable to this article as no new data were created or analyzed in this study.